\documentclass[letterpaper]{article}
\usepackage[preprint]{aaai2027}
\usepackage[hyphens]{url}
\usepackage{graphicx}
\usepackage{natbib}
\usepackage{caption}
\usepackage{amsmath,amssymb,bm}
\usepackage{booktabs}
\usepackage{microtype}
\usepackage{algorithm}
\usepackage{algorithmic}
\usepackage{array}
\usepackage{placeins}
\title{T$^2$exture: Sparsely Perturbed Thermal Texture Imaging\thanks{Preprint version.}\thanks{The code is available at \protect\url{https://github.com/dccc2025/T2exture}, and the data is at \protect\url{https://github.com/JiashuoCHEN/TT-dataset}.}}

\author{
    Jiashuo Chen\equalcontrib\textsuperscript{\rm 1},
    Cheng Dai\equalcontrib\textsuperscript{\rm 2},
    Yanan Hu\textsuperscript{\rm 1},
    Fanglin Bao\textsuperscript{\rm 2}
}
\affiliations{
    \textsuperscript{\rm 1}School of Engineering, Westlake University\\
    \textsuperscript{\rm 2}School of Science, Westlake University\\
    \{chenjiashuo,daicheng,huyanan,baofanglin\}@westlake.edu.cn
}

\newcommand{\texture}{X}
\newcommand{\xhat}{\hat{X}}

\begin{document}
\maketitle

\begin{abstract}
Thermal imaging is effective under adverse illumination, yet passive
long-wave infrared (LWIR) measurements often lack fine texture. Existing thermal texture
imaging approaches commonly rely on spectral sensing or registered auxiliary
modalities, incurring substantial data throughput or vulnerability to
cross-modal degradation. We introduce \textit{T$^2$exture}, a sparsely perturbed thermal texture imaging
framework that aims to reconstruct temporally dense thermal texture sequences from
densely sampled passive frames and a few actively perturbed keyframes. We
define thermal texture as the residual between a source-on observation and its
corresponding source-off passive state. Under sparse LWIR illumination and
rapid quasi-steady paired acquisition, this residual attenuates the
passive-emission background and approximates a source-induced reflected
response, exposing localized material- and geometry-dependent texture.
\textit{T$^2$exture} reconstructs a dense sequence of this source-conditioned
response through two stages. Stage~1 estimates the unobserved source-off
passive state at each active instant from neighboring passive frames to obtain
reliable differential texture anchors. Stage~2 combines sparse anchors with
passive structural context near each target time to reconstruct the dense
sequence. On the simulated benchmark, \textit{T$^2$exture} adds only 0.20M
parameters to AMT-L while improving PSNR by 6.66~dB. Extensive evaluations on
simulated and real acquisitions further show clearer texture recovery and
stronger structural preservation than representative VFI baselines. These
results establish \textit{T$^2$exture} as a practical framework for thermal
texture imaging under sparse active acquisition.
\end{abstract}

\section{Introduction}

Thermal infrared imaging remains informative when visible sensing fails,
enabling applications in autonomous driving
\cite{bao2023heat,ng2024thermalvoyager,zhang2024trafficnight}, environmental
monitoring \cite{aveni2024tirvolch,teng2024sdgsat1}, and medical diagnosis
\cite{han2025phasor,liu2024radicular}. Extracting fine-grained appearance from
passive long-wave infrared (LWIR) imagery, however, remains difficult. A
passive measurement mixes surface self-emission with reflected environmental
radiance. Consequently, distinct combinations of emissivity, temperature, and
environmental radiance can produce similar observations---a phenomenon termed
TeX-degeneracy \cite{bao2023heat,bao2024blurry}. This ambiguity obscures
material- and geometry-dependent appearance cues, limiting fine-grained
perception and downstream recognition.

\begin{figure}[!t]
  \centering
  \includegraphics[width=\linewidth]{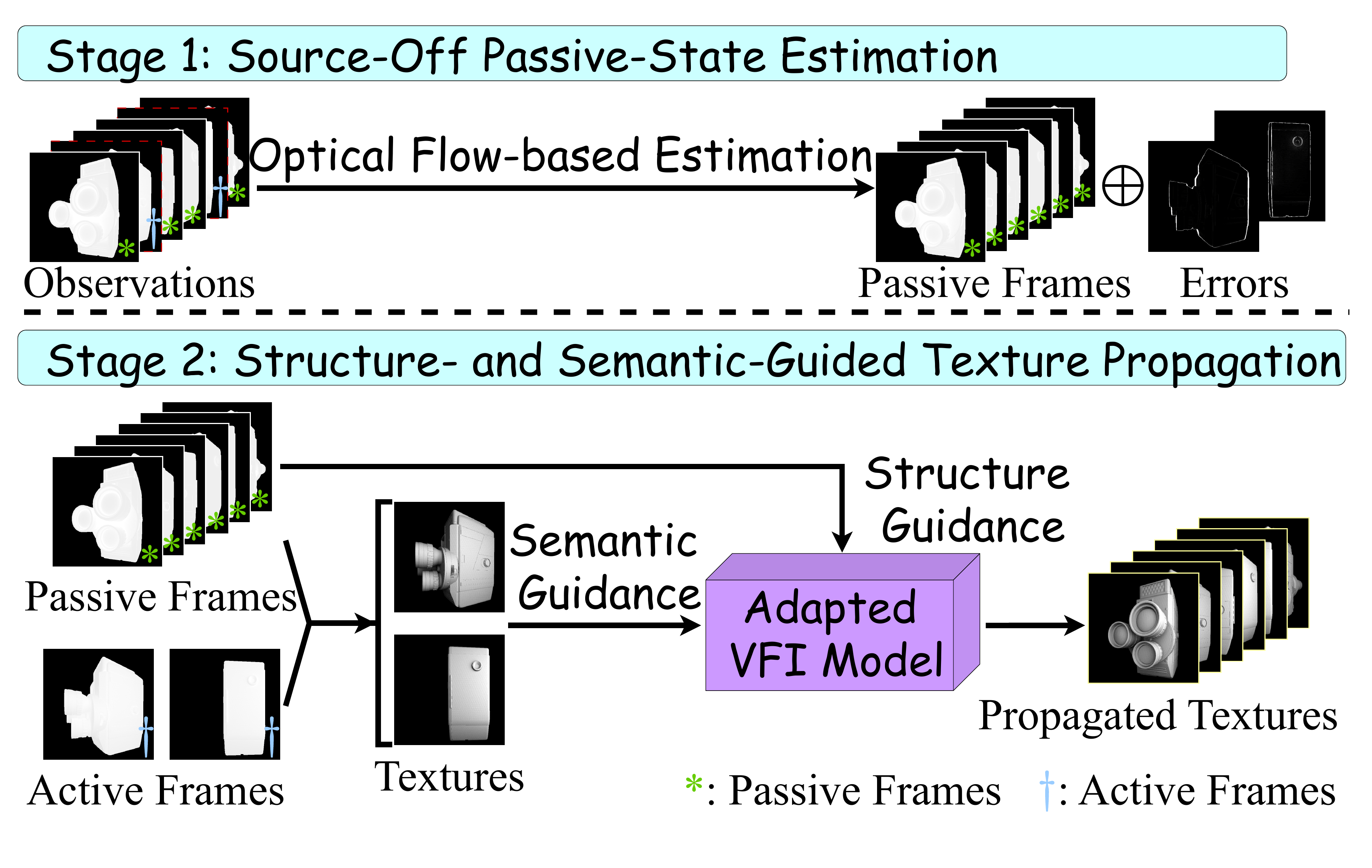}
  \caption{\textbf{Schematic overview of \textit{T$^2$exture}.} Stage~1
  estimates each active keyframe's source-off passive state to obtain sparse texture anchors.
  Stage~2 combines these anchors with
  passive-frame structure in an adapted video frame interpolation (VFI) model
  to reconstruct the sequence.}
  \label{fig:t2exture-overview}
\end{figure}

Existing approaches address this ambiguity by enhancing passive imagery or
relying on additional spectral or auxiliary observations. Image enhancement improves contrast or
sharpens boundaries \cite{zuiderveld1994clahe,liu2019adaptive,hu2024tesr}, but
does not provide the source-conditioned thermal texture evidence absent from
passive observations. Physics-based methods use spectral measurements to
estimate latent temperature--emissivity--texture factors
\cite{bao2023heat,gallastegi2025absorption,gallastegi2026ozone,xu2026tag,
dai2026hair,dai2026tex1500}, at the cost of calibrated acquisition and
substantial data throughput. Visible--thermal fusion transfers appearance from
a registered visible camera \cite{ddfm2023,defusion2025,maskdifuser2026}, but
requires an additional sensor and can degrade under calibration, viewpoint, or
synchronization errors. Active thermal illumination provides a complementary
observation mechanism, yet has primarily targeted inspection or geometric
perception rather than temporally dense reconstruction from sparse active
observations.

\begin{figure}[!t]
  \centering
  \includegraphics[width=\linewidth]{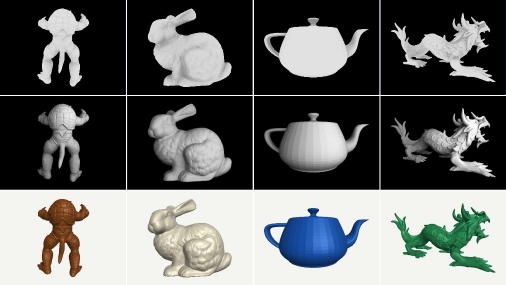}
  \caption{Four samples from our dataset. Rows show thermal
  images, thermal textures (defined in Section~\ref{sec:thermal-texture}), and
  RGB visualizations rendered from surface normals, respectively. }
  \label{fig:simulated-samples}
\end{figure}

We address this gap by proposing \textit{T$^2$exture}, which casts sparsely perturbed
thermal texture imaging as the recovery of a temporally dense,
source-conditioned response sequence from dense passive observations and a few
actively illuminated keyframes. A controlled LWIR source is activated at only
these keyframes, while the remainder of the sequence is captured passively. At
source-visible locations, the difference between a source-on observation and
its corresponding source-off passive state attenuates the passive-emission
background and yields localized, source-conditioned thermal evidence.
Figure~\ref{fig:simulated-samples} visualizes this target: the resulting
thermal-texture maps reveal local appearance cues.
To reconstruct a temporally dense sequence from sparse
evidence, Stage~1 estimates the unobserved source-off passive state at each
active keyframe from neighboring passive frames and forms differential texture
anchors. Stage~2 combines sparse anchors with target-time passive structural
context to reconstruct the dense sequence. We adapt a pretrained video frame interpolation (VFI) model
as a structural and temporal prior for this reconstruction.

Our main contributions are as follows:
\begin{itemize}
  \item We propose \textit{T$^2$exture}, which casts sparsely perturbed thermal
  texture imaging as the recovery of a dense, source-conditioned
  response sequence from dense passive observations and a few actively
  perturbed keyframes.
  \item We define thermal texture as a source-on/source-off radiometric
  residual. Under rapid quasi-steady acquisition, it attenuates the
  passive-emission background and provides localized appearance evidence where
  the controlled source contributes appreciable radiance.
  \item We introduce a two-stage reconstruction pipeline: Stage~1 estimates the
  unobserved source-off passive state from neighboring passive frames to obtain
  differential texture anchors; Stage~2 combines sparse anchors with
  target-time passive structure and an adapted VFI prior for dense temporal
  reconstruction.
  \item We evaluate \textit{T$^2$exture} on simulated and real acquisitions.
  On the simulated benchmark, it adds 0.20M parameters to AMT-L while improving
  PSNR by 6.66~dB; additional experiments assess temporal consistency,
  structural preservation, and robustness under sparse active illumination.
\end{itemize}

\section{Related Work}

\subsubsection{Thermal Texture Recovery}
\label{sec:related-passive-auxiliary}

A passive LWIR observation mixes self-emission and environmental reflection.
Consequently, TeX-degeneracy makes thermal
texture difficult to recover from passive observations alone
\cite{bao2024blurry}.

Approaches that seek physically grounded thermal texture recovery from passive
observations typically introduce additional spectral measurements or an
auxiliary modality.
Hyperspectral approaches fit bandwise radiance with radiative-transfer models,
emissivity priors, and low-rank or spatial regularization to recover latent
thermophysical variables
\cite{bao2023heat,xu2026tag,dai2026hair,dai2026tex1500,liu2026ader,
gallastegi2025absorption,gallastegi2026ozone}. However, they often require
calibrated hyperspectral acquisition and substantial data throughput.

Visible--thermal fusion instead incorporates appearance from a co-registered
visible image while retaining infrared target saliency. Recent
feature-decomposition, task-aware, and diffusion-based fusion methods improve
the fidelity of such fused outputs \cite{cddfuse2023,metafusion2023,ddfm2023,
defusion2025,maskdifuser2026}. Their dependence on a separate sensor makes
them vulnerable to calibration, viewpoint, synchronization, and cross-modal
degradation.

\subsubsection{Active Thermal Illumination}
\label{sec:related-active}

Prior methods introduce active thermal illumination to generate contrast,
commonly through heating, thermal diffusion, or projected thermal patterns.
They have primarily supported nondestructive inspection and geometric
perception: structured LWIR and thermal fringe projection recover 3D shape
\cite{erdozain2020structured,landmann2021thermal,speck2026thermal}, and
laser-painted heat patterns improve correspondence for optical flow, tracking,
and structure from motion \cite{sheinin2024thermal}. These methods enhance
contrast for downstream tasks rather than reconstructing source-conditioned
thermal texture over time.

\subsubsection{Texture Sequence Reconstruction}
\label{sec:related-sparse-reconstruction}

Texture sequence reconstruction is related to video frame interpolation (VFI),
which synthesizes a missing frame from its neighboring observations. Existing
VFI methods broadly follow flow-based, kernel-based, or diffusion-based
formulations: they estimate correspondence and warp inputs, predict
spatially-adaptive resampling kernels, or generate an intermediate latent state,
respectively \cite{li2023amt,liu2024sgmvfi,niklaus2017sepconv,hai2025hfd,
zhang2025eden,lyu2025tlbvfi,peng2026ldfvfi}. Although recent models improve
motion modeling and temporal coherence, long temporal gaps remain challenging.
Without explicit structural conditioning from the dense passive observations
available in our setting, their intermediate reconstructions can suffer
structural instability and texture degradation.

\section{Source-Conditioned Thermal Texture}

This section defines the source-conditioned thermal texture recovered by
\textit{T$^2$exture}. A rapid source-on/source-off acquisition separates
source-induced appearance variation from the slowly varying passive
background, enabling a texture representation from sparse active observations.

\subsubsection{Thermal Image Formation}

We consider a quasi-steady thermal regime in which the surface temperature
field, atmospheric state, and scene geometry vary negligibly over the
acquisition interval. Under this assumption, thermal imaging combines
surface-leaving radiance with atmospheric path radiance
\cite{bao2023heat,gallastegi2025absorption,gallastegi2026ozone,liu2026ader}.
Let $\alpha$ index a surface element, $\nu$ denote wavenumber, and $d_{\alpha}$
be object--camera range. The camera-received spectral radiance is
\begin{equation}
S_{\alpha\nu}
=
\tau_{m\nu}(d_{\alpha})\,S^{\mathrm{surf}}_{\alpha\nu}
+ \left[1-\tau_{m\nu}(d_{\alpha})\right]B_{\nu}(T_m),
\label{eq:camera-path-rte}
\end{equation}
where $S^{\mathrm{surf}}_{\alpha\nu}$ is the radiance leaving the surface,
$\tau_{m\nu}(d_{\alpha})$ is atmospheric transmittance, $B_{\nu}(\cdot)$ is
Planck spectral radiance, and $T_m$ is the effective path-atmosphere
temperature.

Most long-wave infrared cameras operate over $8$--$12\,\mu\mathrm{m}$, where
the atmospheric attenuation coefficient is small, leading to
$\tau_{m\nu}(d_{\alpha})\approx1$ \cite{gallastegi2025absorption}. The
corresponding path-emission term is negligible, so
Eq.~\eqref{eq:camera-path-rte} reduces to
$S_{\alpha\nu}=S^{\mathrm{surf}}_{\alpha\nu}$. For an opaque Lambertian
surface in local thermal equilibrium,
\begin{equation}
S_{\alpha\nu}
= e_{\alpha\nu}B_{\nu}(T_{\alpha})
+ (1-e_{\alpha\nu})\,{E}_{\alpha\nu},
\label{eq:surface-leaving-radiance}
\end{equation}
where $T_{\alpha}$, $e_{\alpha\nu}$, and ${E}_{\alpha\nu}$ are surface
temperature, spectral emissivity, and reflected environmental radiance.
Reflection remains angular even for a Lambertian surface:
\begin{align}
{E}_{\alpha\nu}
&= \frac{1}{\pi}\int_{\Omega_{\alpha}}
S_{\beta\nu}(\omega)\cos\theta\,\mathrm{d}\omega
\notag\\
&= \frac{1}{\pi}\int_{0}^{2\pi}\int_{0}^{\pi/2}
S_{\beta\nu}(\theta,\phi)\cos\theta\sin\theta
\,\mathrm{d}\theta\,\mathrm{d}\phi,
\label{eq:angular-environment-radiance}
\end{align}
Here, $\Omega_{\alpha}$ is the visible hemisphere, $\omega$ is a solid-angle
direction, and $\beta$ indexes the element visible along it. The polar angle
$\theta$ is measured from the normal at $\alpha$; $\phi$ is its azimuth.

\subsubsection{Thermal Texture Definition}
\label{sec:thermal-texture}

Thermal cameras record scalar measurements formed by integrating spectral
radiance over the sensor band against the calibrated response measure
$\mathrm{d}\mu(\nu)$, with $\nu\in[\nu_{\min},\nu_{\max}]$. Let $\Omega_1$
be the passive visible hemisphere and $\Omega_{2,\alpha}\subseteq\Omega_1$
the source-visible domain. The controlled source replaces the environmental
radiance $S_{\beta\nu}$ over $\Omega_{2,\alpha}$ with its blackbody radiance
$L^{\mathrm{src}}_{\alpha\nu}$, giving
\begin{equation}
\begin{aligned}
S^{\mathrm{off}}_{\alpha}
={}& \int_{\nu_{\min}}^{\nu_{\max}}\!\bigg[
e_{\alpha\nu}B_{\nu}(T_{\alpha}) \\
&\qquad + \frac{1-e_{\alpha\nu}}{\pi}\int_{\Omega_1}
S_{\beta\nu}(\omega)\cos\theta\,\mathrm{d}\omega\bigg]
\mathrm{d}\mu(\nu), \\
S^{\mathrm{on}}_{\alpha}
={}& S^{\mathrm{off}}_{\alpha}+\int_{\nu_{\min}}^{\nu_{\max}}
\frac{1-e_{\alpha\nu}}{\pi}\int_{\Omega_{2,\alpha}}\!
\left[L^{\mathrm{src}}_{\alpha\nu}(\omega)-S_{\beta\nu}(\omega)\right] \\
&\qquad\qquad\qquad\qquad\cos\theta\,\mathrm{d}\omega\,\mathrm{d}\mu(\nu).
\end{aligned}
\label{eq:band-integrated-on-off}
\end{equation}
We define source-conditioned thermal texture as the nonnegative paired
residual
\begin{equation}
X_{\alpha}\triangleq\left[S_{\alpha}^{\mathrm{on}}
-S_{\alpha}^{\mathrm{off}}\right]_{+},
\qquad [z]_{+}=\max(0,z).
\label{eq:thermal-texture-residual}
\end{equation}
Under rapid quasi-steady paired acquisition, $X_{\alpha}$ approximates a
source-induced reflected response. Its spatial variation exposes localized
material- and geometry-dependent texture where the controlled source
contributes appreciable radiance. The nonnegative projection suppresses weak
negative differences caused by noise or residual temporal mismatch. Thus,
thermal texture is not a universal surface measurement; it is a localized,
source-conditioned appearance representation.

\subsubsection{Task Formulation}

We consider a thermal camera observing a dynamic scene over $N$ frames. Most
frames are acquired passively, while active illumination is applied at sparse
keyframes. Let $\mathcal{K}\subset\{1,\ldots,N\}$ index the active keyframes
and $\overline{\mathcal{K}}=\{1,\ldots,N\}\setminus\mathcal{K}$ the passive
ones. The observations comprise passive frames
$\mathcal{S}^{\mathrm{off}}=\{S_i^{\mathrm{off}}\}_{i\in\overline{\mathcal K}}$
and active frames
$\mathcal{S}^{\mathrm{on}}=\{S_k^{\mathrm{on}}\}_{k\in\mathcal K}$. At an active keyframe, the corresponding source-off passive state is
unobserved. Given the mixed observations, the task is to recover the
temporally dense source-conditioned thermal-texture sequence
$\mathcal X=\{X_i\}_{i=1}^{N}$:
\begin{equation}
\hat{\mathcal{X}}=f_\theta\!\left(
\mathcal{S}^{\mathrm{off}},
\mathcal{S}^{\mathrm{on}}\right).
\end{equation}
This formulation motivates texture-sequence propagation in which sparse active
frames provide texture evidence and dense passive frames provide target-time
structural context.

\section{Method}

\begin{figure*}[!t]
  \centering
  \includegraphics[width=\textwidth]{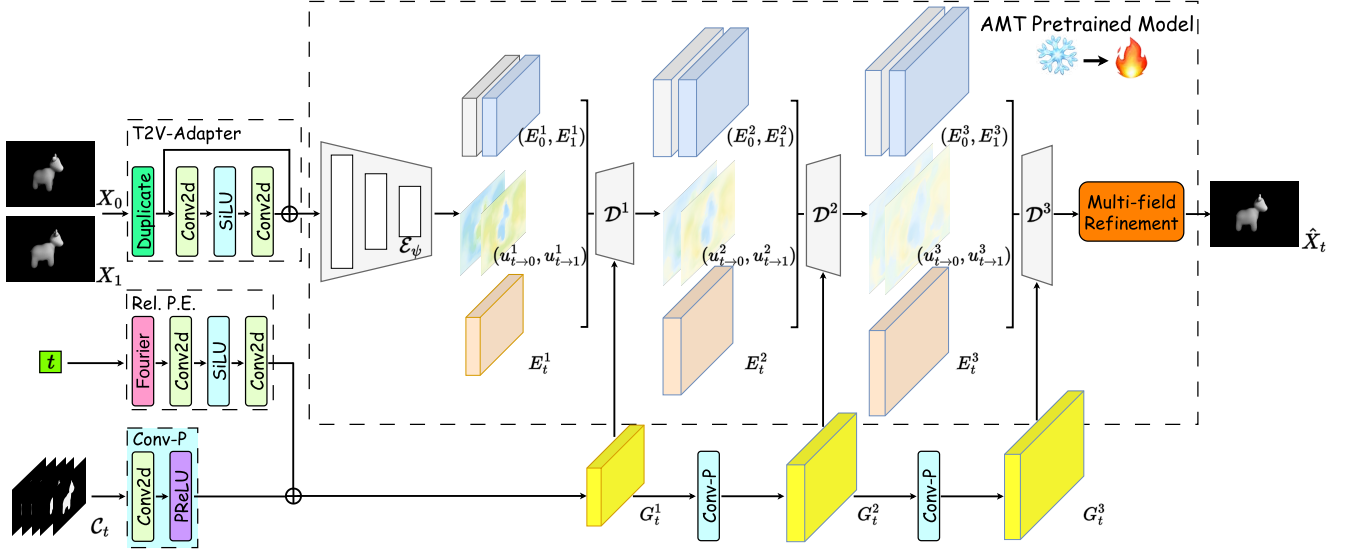}
  \caption{\textbf{Illustration of T$^2$exture second Stage framework.}
The T2V-Adapter maps texture anchors to the AMT feature space~\cite{li2023amt},
while convolutional adapters inject multi-scale passive context into its decoder
to synthesize $\hat{X}_t$.}
  \label{fig:stage2-architecture}
\end{figure*}

\subsection{Overview}
\label{sec:method-overview}

T$^2$exture reconstructs a dense sequence of source-conditioned thermal textures
from sparse active measurements and dense passive observations. As illustrated
in Fig.~\ref{fig:stage2-architecture}, it instantiates this formulation with
two VFI-based stages. Stage~1 uses a pretrained VFI model to estimate the
source-off passive state at each active keyframe, yielding source-conditioned
texture anchors. Stage~2 employs an adapted VFI model, queried at normalized
time $t\in(0,1)$, to propagate texture between anchor pairs while conditioning
on nearby passive frames as target-time structural context.

\subsection{Source-Off Passive-State Estimation}
\label{sec:method-anchors}

At an active keyframe $k$, the passive state $S_k^{\mathrm{off}}$ is
unobserved. We instantiate $f_\theta$ with the pretrained AMT video frame
interpolation model~\cite{li2023amt} and estimate this source-off state from
its two adjacent passive observations:
\begin{equation}
\widehat{S}_k^{\mathrm{off}} = f_\theta\!\left(S_{k-1}^{\mathrm{off}},
S_{k+1}^{\mathrm{off}}, t=\dfrac{1}{2}\right).
\label{eq:source-off-estimation}
\end{equation}
The texture anchor is then defined based on Eq.~\eqref{eq:thermal-texture-residual}:
\begin{equation}
X_k=\left[S_k^{\mathrm{on}}-\widehat{S}_k^{\mathrm{off}}\right]_+,
\label{eq:texture-anchor}
\end{equation}
For an interval bracketed by two active keyframes, the resulting anchors are
denoted by $X_0$ and $X_1$.

\subsection{Structure-Semantic-Guided Texture Propagation}
\label{sec:method-propagation}

\subsubsection{Texture-to-Visual Adaptation}
\label{sec:method-t2v}

AMT is pretrained on natural video, whereas a thermal texture is a
single-channel residual with a different numerical distribution.
We therefore introduce a lightweight texture-to-visual (T2V) adapter
$\mathcal{A}_\phi$ before the pretrained AMT encoder. The adapter first
replicates the texture channel and then refines it using two convolutional
layers separated by a SiLU nonlinearity:
\begin{equation}
\widetilde{X}_i=\mathcal{A}_\phi(X_i)\in\mathbb{R}^{3\times H\times W},
\qquad i\in\{0,1\}.
\label{eq:t2v-adapter}
\end{equation}
$\mathcal{A}_\phi$ learns a representation suitable for the
pretrained AMT model, preserving the motion
and interpolation prior of AMT while allowing it to operate on thermal-texture
anchors.

\subsubsection{Temporal Passive Structural Guidance}
\label{sec:method-passive-guidance}

Texture anchors expose source-induced appearance only at sparse active
keyframes; reconstruction from anchors alone is therefore structurally
underconstrained. Stage~1 completes the source-off sequence by inserting its
estimates at active keyframes: we denote the resulting length-$N$ sequence by
$\widetilde{\mathcal{S}}^{\mathrm{off}}=
\{\widetilde{S}_i^{\mathrm{off}}\}_{i=1}^{N}$, where
$\widetilde{S}_i^{\mathrm{off}}=S_i^{\mathrm{off}}$ for
$i\in\overline{\mathcal{K}}$ and
$\widetilde{S}_i^{\mathrm{off}}=\widehat{S}_i^{\mathrm{off}}$ for
$i\in\mathcal{K}$. For a target time
$t$, we extract a context from the sequence:
\begin{equation}
\begin{aligned}
\mathcal{C}_t={}&\Big(\big(\widetilde{S}_{\tau_i^-}^{\mathrm{off}}\big)_{i=r}^{1},
\widetilde S_t^{\mathrm{off}},
\big(\widetilde{S}_{\tau_i^+}^{\mathrm{off}}\big)_{i=1}^{r}\Big),\\
&\tau_r^-<\cdots<\tau_1^-<t<\tau_1^+<\cdots<\tau_r^+.
\end{aligned}
\label{eq:passive-context}
\end{equation}
This complete source-off context supplies structural guidance for
texture propagation. To specify the reconstruction time, we encode $t$ with
Fourier features~\cite{tancik2020fourier}:
\begin{equation}
\gamma(t)=\left[\sin(2\pi f_kt),\cos(2\pi f_kt)\right]_{k=1}^{K},
\end{equation}
which a two-layer SiLU MLP maps to temporal modulation
$c_t=\mathcal{M}(\gamma(t))$. Conditioned on $c_t$, the passive encoder
constructs a three-scale structural pyramid:
\begin{equation}
\begin{aligned}
G_t^1 &= \mathcal{P}_1(\mathcal{C}_t)+c_t,\\
G_t^{l+1} &= \mathcal{P}_{l+1}(G_t^l),\quad l\in\{1,2\}.
\end{aligned}
\label{eq:passive-pyramid}
\end{equation}
The resulting $\{G_t^1,G_t^2,G_t^3\}$ encode target-time passive structure at the
three AMT decoder resolutions. Following~\cite{mou2024t2iadapter},
we inject each feature into its corresponding decoder stage by residual adaptation:
\begin{equation}
\overline{F}^{l}=F^{l}+\mathcal{Z}_{l}(G_t^{l}),
\label{eq:residual-injection}
\end{equation}
where $\mathcal{Z}_{l}$ is a zero-initialized $1\times1$ convolution, preserving
the pretrained backbone at initialization while enabling structural adaptation
during fine-tuning. AMT's multi-field refinement aggregates flow-based candidates
from the conditioned decoder to yield $\widehat{X}_t$.

\subsection{Loss Functions}
\label{sec:method-training}

Following the loss design of AMT~\cite{li2023amt}, we combine a Charbonnier
reconstruction loss~\cite{charbonnier1994deterministic}, a bidirectional
census loss~\cite{meister2018unflow}, and the multi-scale flow-distillation
loss of IFRNet~\cite{kong2022ifrnet}:
\begin{equation}
\mathcal{L}=\lambda_{\mathrm{char}}\mathcal{L}_{\mathrm{char}}
+\lambda_{\mathrm{css}}\mathcal{L}_{\mathrm{css}}
+\lambda_{\mathrm{flow}}\mathcal{L}_{\mathrm{flow}}.
\label{eq:training-objective}
\end{equation}
The Charbonnier term preserves per-pixel texture fidelity, while the census
term preserves local structure and boundary consistency. The flow term constrains intermediate multi-scale bilateral flow
predictions. We empirically set
$(\lambda_{\mathrm{char}},\lambda_{\mathrm{css}},\lambda_{\mathrm{flow}})
=(1.0,0.1,10^{-3})$, prioritizing consistent texture synthesis compared with AMT~\cite{li2023amt}.

\section{Experiments}
\label{sec:experiments}

\subsection{Experiment Setup}
\label{sec:implementation-details}

\subsubsection{Dataset Construction}
To match the intended sparse active LWIR setting, we construct paired
simulated and real-world datasets for fine-tuning and evaluation. The simulated
set contains 32 objects drawn from Common 3D Test Models
and Poly Haven, each rendered as 180
paired active/passive frames at \(960\times640\) resolution. We use an
object-disjoint 20/4/8 split for training, validation, and testing. Active
keyframes are sampled every ten frames, and the nine intervening times are
reconstruction targets with dense passive observations. Thermal radiance
transport is rendered with an extended spectral renderer. The
real-world set comprises six captured active/passive LWIR sequences at
\(1024\times1280\) resolution and is used only for inference because dense
texture ground truth is unavailable. Active frames in both datasets are formed
under rapid blackbody illumination. Full
acquisition details are provided in the supplement.

\subsubsection{Evaluation Metrics}
For the simulated benchmark with texture ground truth, we report PSNR, SSIM,
IE, NIE, and Edge-F1@2px. PSNR, SSIM, IE, and NIE quantify pixel-domain
fidelity and interpolation accuracy, whereas Edge-F1@2px measures boundary
preservation by matching Canny edge maps within a two-pixel tolerance
\cite{canny1986computational,arbelaez2011contour}. For real sequences without
dense ground truth, we report the no-reference metrics En, AG, SD, and SCD
to characterize information content, contrast, and spatial detail
\cite{aslantas2015scd}, together with PI for perceptual image quality
\cite{wang2024multifocus}. Detailed definitions are
provided in the supplement.

\subsubsection{Implementation Details}
We implement \textit{T$^2$exture} in PyTorch~2.6.0 with CUDA~12.4 on a single
NVIDIA RTX~4090 GPU. Both stages use the S, L, and G AMT backbones.
Stage~1 applies the corresponding pretrained backbone only
for source-off-state inference, whereas Stage~2 initializes from the matching
checkpoint and adapts it for texture propagation. Stage~2 training uses AdamW
with $(\beta_1,\beta_2)=(0.9,0.99)$ and weight decay $10^{-5}$. We first train
its adapters for 10,000 iterations with learning rate $2\times10^{-4}$, then
jointly fine-tune all Stage~2 modules for 5,000 iterations with learning rate
$5\times10^{-5}$. Training samples are randomly cropped to \(384\times384\)
without additional augmentation; testing uses full-resolution frames. Unless
otherwise stated, Ours-L is the default configuration and sets $r=2$ in
$\mathcal{C}_t$, combining the target-time passive observation
$S_t^{\mathrm{off}}$ with two neighboring source-off states on either side.
We train on the simulated split and apply the resulting model to both simulated
and real-world benchmarks. Additional experimental details are provided in the
supplement.

\subsection{Comparison with Prior Work}
\label{sec:prior-work-comparison}

\subsubsection{Comparison Methods}
For Stage~1 source-off-state estimation, we compare RAFT \cite{teed2020raft} and GMA
\cite{jiang2021gma}, representative optical-flow estimators used to reconstruct
the intermediate passive state, with AMT-L \cite{li2023amt}. For Stage~2,
we compare \textit{T$^2$exture} with representative VFI baselines, including
IFRNet \cite{kong2022ifrnet}, SGM-VFI \cite{liu2024sgmvfi}, BiM-VFI
\cite{seo2025bimvfi}, GIMM-F \cite{guo2024gimmvfi}, and AMT-L.

\subsubsection{Quantitative and Qualitative Comparisons}

\begin{figure}[!t]
  \centering
  \includegraphics[width=\columnwidth]{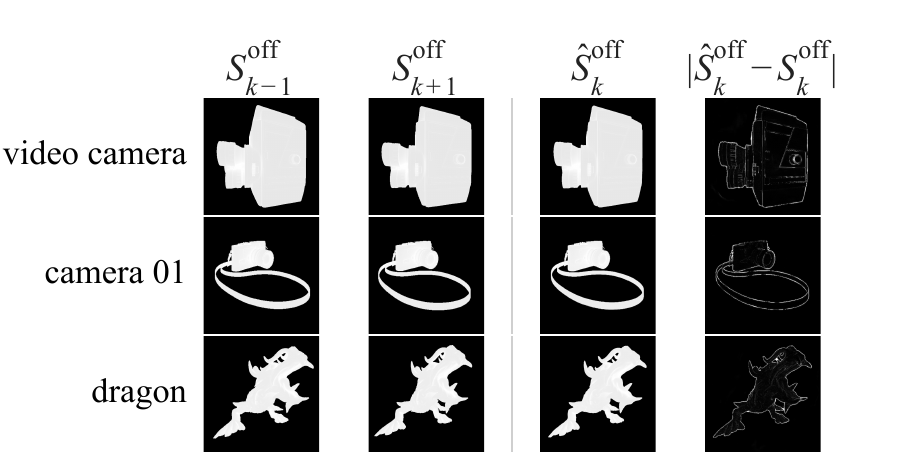}
  \caption{\textbf{Stage~1 source-off-state prediction.} From adjacent
passive frames, AMT-L predicts the intermediate source-off passive state; the
rightmost column shows the error.}
  \label{fig:source-off-qualitative}
\end{figure}

For Stage~1, Table~\ref{tab:source-off-results} and
Fig.~\ref{fig:source-off-qualitative} show that AMT-L accurately estimates the
intermediate source-off state, preserving its structure with only minor motion estimation
error.

\begin{table}[!t]
\centering
\small
\caption{\textbf{Stage~1 source-off passive-state estimation on the simulated
benchmark.} Best results are in bold.}
\label{tab:source-off-results}
\setlength{\tabcolsep}{2.2pt}
\renewcommand{\arraystretch}{1.15}
\begin{tabular}{@{}lcccccc@{}}
\toprule
\textbf{Method} & \textbf{Params} & \textbf{PSNR}$\uparrow$ & \textbf{SSIM}$\uparrow$ & \textbf{E-F1}$\uparrow$ & \textbf{IE}$\downarrow$ & \textbf{NIE}$\downarrow$ \\
\midrule
RAFT & 5.30 & 32.227 & 0.992 & 0.965 & 0.482 & 0.002 \\
GMA & 5.90 & 32.495 & 0.992 & 0.971 & 0.466 & 0.002 \\
AMT-L & 12.94 & \textbf{35.237} & \textbf{0.996} & \textbf{0.995} & \textbf{0.240} & \textbf{0.001} \\
\bottomrule
\multicolumn{7}{@{}l}{\footnotesize \textit{Units:} Params(M), PSNR (dB).} \\
\end{tabular}
\end{table}

\begin{figure*}[!t]
  \centering
  \includegraphics[width=\textwidth]{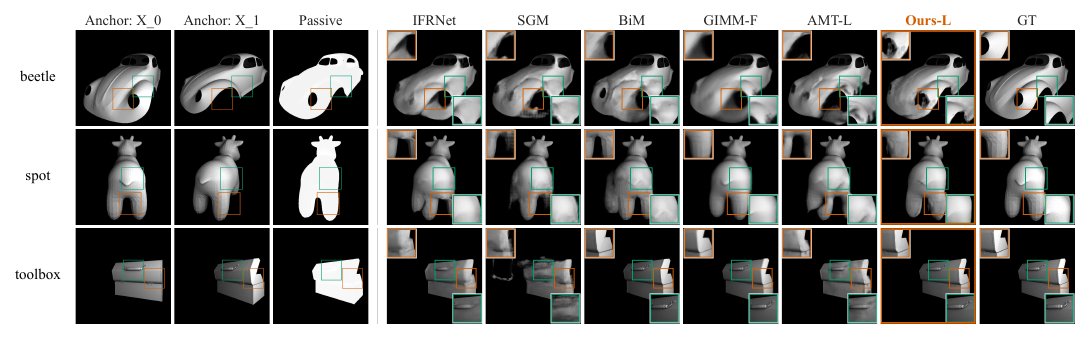}
  \caption{\textbf{Qualitative reconstruction comparison on the simulated
  benchmark.} Each row shows sparse anchors, the target-time passive observation,
  ground truth, prior VFI results, and Ours-L. Insets enlarge selected regions of interest.}
  \label{fig:qualitative-reconstruction}
\end{figure*}

For Stage~2, Table~\ref{tab:simulated-results} reports results on the
simulated benchmark. With 0.20M additional parameters over AMT-L, Ours-L
improves PSNR by 6.66~dB and Edge-F1 by 0.018 while achieving the best SSIM,
IE, and NIE. The lightweight adaptation improves texture reconstruction without
sacrificing the efficiency of the pretrained backbone. Although BiM-VFI attains
slightly higher Edge-F1, it incurs larger reconstruction errors.
Figure~\ref{fig:qualitative-reconstruction} reveals its missing structures and
artifacts in important regions, whereas Ours-L better preserves boundaries,
thin structures, and texture alignment.

\begin{table}[!t]
\centering
\small
\caption{\textbf{Quantitative comparison on the simulated benchmark.} Best and
second-best results are shown in bold and underlined, respectively.}
\label{tab:simulated-results}
\setlength{\tabcolsep}{2.1pt}
\renewcommand{\arraystretch}{1.28}
\begin{tabular}{@{}lcccccc@{}}
\toprule
\textbf{Method} & \textbf{Params} & \textbf{PSNR}$\uparrow$ & \textbf{SSIM}$\uparrow$ & \textbf{E-F1}$\uparrow$ & \textbf{IE}$\downarrow$ & \textbf{NIE}$\downarrow$ \\
\midrule
IFRNet & 5.00 & 25.435 & 0.942 & 0.957 & 3.097 & 0.012 \\
SGM-VFI & 20.80 & 25.442 & 0.941 & 0.951 & 3.170 & 0.012 \\
BiM-VFI & 6.88 & 29.472 & 0.956 & \textbf{0.970} & 1.927 & 0.008 \\
GIMM-F & 30.61 & \underline{31.340} & \underline{0.962} & 0.912 & \underline{1.351} & \underline{0.005} \\
AMT-L & 12.94 & 25.351 & 0.940 & 0.945 & 3.294 & 0.013 \\
Ours-L & 13.14 & \textbf{32.014} & \textbf{0.967} & \underline{0.963} & \textbf{1.157} & \textbf{0.005} \\
\bottomrule
\multicolumn{7}{@{}l}{\footnotesize \textit{Units:} Params(M), PSNR (dB).} \\
\end{tabular}
\end{table}

\begin{figure*}[!t]
  \centering
  \includegraphics[width=\textwidth]{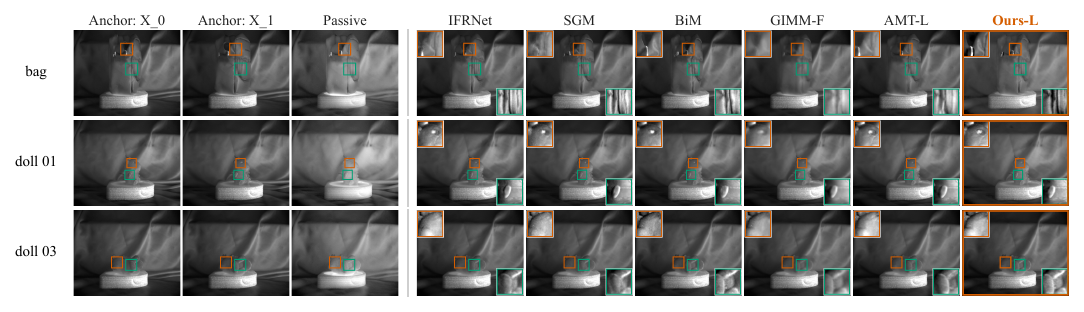}
  \caption{Qualitative reconstruction comparison on real-world sequences. Each
  row shows two sparse active anchors, the passive target-time
  observation, and reconstructions from prior VFI baselines and Ours-L. Insets
  enlarge selected regions of interest.}
  \label{fig:qualitative-real}
\end{figure*}

For real-world sequences, dense texture references are unavailable; accordingly,
Table~\ref{tab:real-results} reports no-reference metrics. Ours-L achieves the
highest entropy (En), standard deviation (SD), and sum of correlations of
differences (SCD), suggesting richer information content, stronger contrast
variation, and more structural detail in real-world reconstructions. Although no
single method dominates every metric, Figure~\ref{fig:qualitative-real}
complements it with visual evidence: on the zipper, doll eyes,
and body texture, Ours-L recovers finer details while more reliably preserving
object structure than competing methods. 

Across simulated and real settings,
these results demonstrate that, under joint structural and semantic guidance,
our framework achieves stable, high-quality texture propagation.

\begin{table}[!t]
\centering
\small
\caption{\textbf{Quantitative comparison on the real-world benchmark.} Best and
second-best results are shown in bold and underlined, respectively.}
\label{tab:real-results}
\setlength{\tabcolsep}{2.1pt}
\renewcommand{\arraystretch}{1.28}
\begin{tabular}{@{}lcccccc@{}}
\toprule
\textbf{Method} & \textbf{Params} & \textbf{En}$\uparrow$ & \textbf{AG}$\uparrow$ & \textbf{SD}$\uparrow$ & \textbf{SCD}$\uparrow$ & \textbf{PI}$\downarrow$ \\
\midrule
IFRNet & 5.00 & 6.574 & 1.382 & 27.899 & 0.005 & 5.516 \\
SGM-VFI & 20.80 & \underline{6.587} & \textbf{1.673} & \underline{27.940} & \underline{0.061} & 5.280 \\
BiM-VFI & 6.88 & 6.582 & \underline{1.585} & 27.930 & 0.015 & \textbf{5.159} \\
GIMM-F & 30.61 & 6.574 & 1.264 & 27.883 & -0.001 & 5.616 \\
AMT-L & 12.94 & 6.577 & 1.502 & 27.911 & 0.018 & \underline{5.175} \\
Ours-L & 13.14 & \textbf{6.621} & 1.193 & \textbf{28.902} & \textbf{0.359} & 5.384 \\
\bottomrule
\multicolumn{7}{@{}l}{\footnotesize \textit{Units:} Params(M).} \\
\end{tabular}
\end{table}

\subsubsection{Efficiency and Scalability}
To assess whether \textit{T$^2$exture} transfers stably and efficiently across
backbone capacities, we fine-tune AMT-S, AMT-L, and AMT-G under a shared
protocol; detailed settings are provided in the supplement. Table~\ref{tab:model-scale-ablation}
shows the adaptation adds only 0.066--0.487M parameters across scales,
with modest overheads of 0.58--1.71~ms in latency and 1.5\%--2.5\% in GFLOPs
per frame. In particular, T$^2$exture-S runs at 21.54~ms per frame
(approximately 46~FPS), supporting real-time texture generation in lightweight
deployment. In return, it consistently improves PSNR by
5.06--7.49~dB and Edge-F1 by 0.013--0.018, indicating improved
texture fidelity and structural preservation. 

These results demonstrate
that our structural and semantic adaptation scales effectively across AMT model
sizes while retaining a favorable accuracy--efficiency trade-off.

\begin{table}[!t]
\centering
\fontsize{7.6}{8.7}\selectfont
\caption{Quantitative comparison with AMT baselines across model scales on the simulated benchmark.}
\label{tab:model-scale-ablation}
\setlength{\tabcolsep}{1.8pt}
\renewcommand{\arraystretch}{1.10}
\begin{tabular}{@{}lcccccccc@{}}
\toprule
\textbf{Method} & \textbf{Params}& \textbf{PSNR}$\uparrow$ & \textbf{SSIM}$\uparrow$ & \textbf{E-F1}$\uparrow$ & \textbf{IE}$\downarrow$ & \textbf{NIE}$\downarrow$ & \textbf{Lat.}$\downarrow$ & \textbf{GFLOPs}$\downarrow$ \\
\midrule
AMT-S  & 2.99  & 24.370 & 0.934 & 0.949 & 3.736 & 0.015 & \textbf{20.955} & \textbf{135.198} \\
Ours-S & 3.06  & \textbf{29.426} & \textbf{0.959} & \textbf{0.962} & \textbf{1.601} & \textbf{0.006} & 21.537 & 138.513 \\
\cmidrule(lr){1-9}
AMT-L  & 12.94 & 25.351 & 0.940 & 0.945 & 3.294 & 0.013 & \textbf{43.191} & \textbf{715.251} \\
Ours-L & 13.14 & \textbf{32.014} & \textbf{0.967} & \textbf{0.963} & \textbf{1.157} & \textbf{0.005} & 44.136 & 729.955 \\
\cmidrule(lr){1-9}
AMT-G  & 30.64 & 25.436 & 0.940 & 0.944 & 3.288 & 0.013 & \textbf{81.855} & \textbf{2640.905} \\
Ours-G & 31.13 & \textbf{32.924} & \textbf{0.970} & \textbf{0.960} & \textbf{1.100} & \textbf{0.004} & 83.566 & 2680.474 \\
\bottomrule
\multicolumn{9}{@{}l@{}}{\scriptsize \textit{Units:} Params (M), PSNR (dB), Lat. (ms).} \\
\end{tabular}
\end{table}

\subsection{Ablation Study}
\label{sec:ablation-study}

We conduct controlled studies on the simulated benchmark to examine the
contributions of our design and the effects of active-illumination sparsity and
passive structural context.

\subsubsection{Component Ablation}
Table~\ref{tab:component-ablation} shows that T2V-Adapter improves PSNR by 3.14~dB over
AMT-L, and passive guidance without temporal modulation adds 2.44~dB. The full
model obtains the highest PSNR and SSIM and the lowest IE and NIE; despite lower
Edge-F1 than the unmodulated variant, it delivers the strongest overall
reconstruction.

\begin{table}[t]
\centering
\small
\caption{\textbf{Component ablation of the T$^2$exture model.}}
\label{tab:component-ablation}
\setlength{\tabcolsep}{2.2pt}
\renewcommand{\arraystretch}{1.12}
\begin{tabular}{@{}lccccc@{}}
\toprule
\textbf{Variant} & \textbf{PSNR}$\uparrow$ & \textbf{SSIM}$\uparrow$ & \textbf{E-F1}$\uparrow$ & \textbf{IE}$\downarrow$ & \textbf{NIE}$\downarrow$ \\
\midrule
AMT-L & 25.3512 & 0.9401 & 0.9448 & 3.2938 & 0.0129 \\
+ T2V & 28.4882 & 0.9553 & 0.9586 & 1.9352 & 0.0076 \\
+ Guidance w/o T.M. & \underline{30.9271} & \underline{0.9640} & \textbf{0.9674} & \underline{1.4406} & \underline{0.0056} \\
Full T$^2$exture & \textbf{32.0136} & \textbf{0.9668} & \underline{0.9626} & \textbf{1.1571} & \textbf{0.0045} \\
\bottomrule
\multicolumn{6}{@{}l}{\scriptsize \textit{Units:} PSNR (dB). T.M.: relative temporal modulation.} \\
\end{tabular}
\end{table}

\subsubsection{Active-Illumination Sparsity}
We vary the interval of passive frames between two active anchors.
 As it increases from 1 to 10, PSNR decreases from
40.10 to 31.40~dB while IE increases from 0.41 to 1.33
(Table~\ref{tab:active-sparsity}). This result exposes the practical
trade-off of sparse active acquisition: wider anchor spacing
reduces active exposure, whereas denser sampling provides
more reliable texture reconstruction.

\begin{table}[t]
\centering
\small
\caption{\textbf{Effect of active-illumination sparsity on the simulated benchmark.}
Interval denotes the number of passive frames between two active anchors.}
\label{tab:active-sparsity}
\setlength{\tabcolsep}{3.0pt}
\renewcommand{\arraystretch}{1.12}
\begin{tabular}{@{}cccccc@{}}
\toprule
\textbf{Interval} & \textbf{PSNR}$\uparrow$ & \textbf{SSIM}$\uparrow$ & \textbf{E-F1}$\uparrow$ & \textbf{IE}$\downarrow$ & \textbf{NIE}$\downarrow$ \\
\midrule
1  & \textbf{40.1030} & \textbf{0.9883} & \textbf{0.9781} & \textbf{0.4100} & \textbf{0.0016} \\
2  & \underline{37.1258} & \underline{0.9826} & \underline{0.9705} & \underline{0.5780} & \underline{0.0023} \\
3  & 35.7422 & 0.9794 & 0.9593 & 0.6732 & 0.0026 \\
4  & 34.2812 & 0.9765 & 0.9566 & 0.8065 & 0.0032 \\
5  & 33.3159 & 0.9726 & 0.9567 & 0.9720 & 0.0038 \\
6  & 32.7287 & 0.9701 & 0.9574 & 1.0809 & 0.0042 \\
7  & 32.5649 & 0.9686 & 0.9604 & 1.0914 & 0.0043 \\
8  & 32.1998 & 0.9679 & 0.9613 & 1.1377 & 0.0045 \\
9  & 32.0136 & 0.9668 & 0.9626 & 1.1571 & 0.0045 \\
10 & 31.4042 & 0.9646 & 0.9640 & 1.3272 & 0.0052 \\
\bottomrule
\multicolumn{6}{@{}l}{\scriptsize \textit{Units:} PSNR (dB).} \\
\end{tabular}
\end{table}

\subsubsection{Passive Structural Context}
We vary the size of the context in Eq.~\eqref{eq:passive-context}. Adding the
central target-time passive observation ($|\mathcal{C}_t|=1$) raises PSNR from 28.49 to
31.91~dB, while a five-frame context reaches 32.01~dB. Performance saturates
for $|\mathcal{C}_t|=5$--$7$ (Table~\ref{tab:passive-context}), confirming
that nearby passive context constrains target-time structure during texture
propagation. More distant frames provide diminishing returns as temporal
correspondence weakens; we therefore use $|\mathcal{C}_t|=5$ by default.

\begin{table}[t]
\centering
\small
\caption{\textbf{Effect of passive structural context on the simulated benchmark.}
$|\mathcal{C}_t|$ denotes the context size in Eq.~\eqref{eq:passive-context};
None omits the context.}
\label{tab:passive-context}
\setlength{\tabcolsep}{3.0pt}
\renewcommand{\arraystretch}{1.12}
\begin{tabular}{@{}cccccc@{}}
\toprule
\textbf{$|\mathcal{C}_t|$} & \textbf{PSNR}$\uparrow$ & \textbf{SSIM}$\uparrow$ & \textbf{E-F1}$\uparrow$ & \textbf{IE}$\downarrow$ & \textbf{NIE}$\downarrow$ \\
\midrule
None  & 28.4882 & 0.9553 & 0.9586 & 1.9352 & 0.0076 \\
1  & 31.9144 & 0.9657 & 0.9618 & 1.2340 & 0.0047 \\
3  & \underline{32.0116} & \underline{0.9669} & 0.9622 & \underline{1.1761} & 0.0046 \\
5  & \textbf{32.0136} & 0.9668 & \underline{0.9626} & \textbf{1.1571} & \textbf{0.0045} \\
7  & 31.8853 & \textbf{0.9671} & \textbf{0.9633} & 1.1834 & \underline{0.0046} \\
9  & 31.7464 & 0.9664 & 0.9618 & 1.2011 & 0.0047 \\
11 & 31.6118 & 0.9663 & 0.9625 & 1.2173 & 0.0048 \\
\bottomrule
\multicolumn{6}{@{}l}{\scriptsize \textit{Units:} PSNR (dB).} \\
\end{tabular}
\end{table}

\subsection{Safety Considerations}
\label{sec:safety-deployability}

Our prototype uses a noncoherent, extended-area blackbody source in the LWIR
band. Safe deployment requires calibrated measurements below the applicable
ICNIRP limits for incoherent infrared radiation, which depend on source radiance
and angular extent, distance, exposure duration, and duty cycle
\cite{icnirp2013incoherent}. Sparse, short-duration keyframes reduce the duty
cycle; in the far field, irradiance from a point-like source decreases
approximately with the inverse square of distance, whereas an extended source
requires direct exposure measurement. A calibrated stand-off distance and
avoidance of contact with the heated source housing mitigate thermal hazards.

\section{Conclusion}
\label{sec:conclusion}

We presented \textit{T$^2$exture}, a sparsely perturbed thermal texture imaging framework that reconstructs
temporally dense source-conditioned thermal texture under TeX-degeneracy. It
defines thermal texture as a source-on/source-off residual that attenuates
passive emission and exposes localized material- and geometry-dependent
evidence. Stage~1 uses pretrained AMT to estimate the source-off passive state
from neighboring passive frames and form differential anchors. Stage~2 adapts
AMT to propagate these anchors under target-time structural constraints supplied
by $\mathcal{C}_t$. Experiments on simulated and real acquisitions show accurate
source-off estimation and improved texture reconstruction and structure
preservation over direct VFI transfer with modest overhead.

\clearpage
\appendix
\setcounter{section}{0}
\setcounter{figure}{0}
\setcounter{table}{0}
\setcounter{equation}{0}
\setcounter{algorithm}{0}
\renewcommand{\thesection}{S\arabic{section}}
\renewcommand{\thesubsection}{\thesection.\arabic{subsection}}
\renewcommand{\thetable}{S\arabic{table}}
\renewcommand{\thefigure}{S\arabic{figure}}
\renewcommand{\theequation}{S\arabic{equation}}
\renewcommand{\thealgorithm}{S\arabic{algorithm}}
\section*{Technical Supplementary Material}
\label{sec:supplementary-material}

\section{Dataset Details}
\label{sec:dataset}

We construct complementary synthetic and real LWIR benchmarks for controlled
evaluation and real-world transfer assessment. The synthetic benchmark provides
paired source-on and source-off observations with dense texture targets, and is
used for training, model selection, and full-reference evaluation. The real
benchmark consists of independently captured LWIR sequences and is used only
for inference-time, no-reference evaluation, which does
not provide dense texture ground truth.

\subsection{Synthetic Benchmark}

\paragraph{Dataset overview.}
The synthetic benchmark contains 32 fixed object sequences, each with a full
sequence of 180 ordered views at \(960\times640\) resolution. Default baseline
comparisons use the shared 170-view subset specified in the main paper. Paired
source-on and source-off renders share the same object pose, camera, and scene
geometry; only the active source state differs. The source-conditioned texture is
\(X=[S^{\mathrm{on}}-S^{\mathrm{off}}]_+\), where \([u]_+=\max(u,0)\).

The split is object-disjoint, with 20, 4, and 8 scenes for training,
validation, and testing, respectively. Active anchors are spaced ten views
apart; the nine intervening views serve as targets, yielding 152 target-time
samples per scene. 

\begin{figure*}[!t]
\centering
\includegraphics[width=0.99\textwidth]{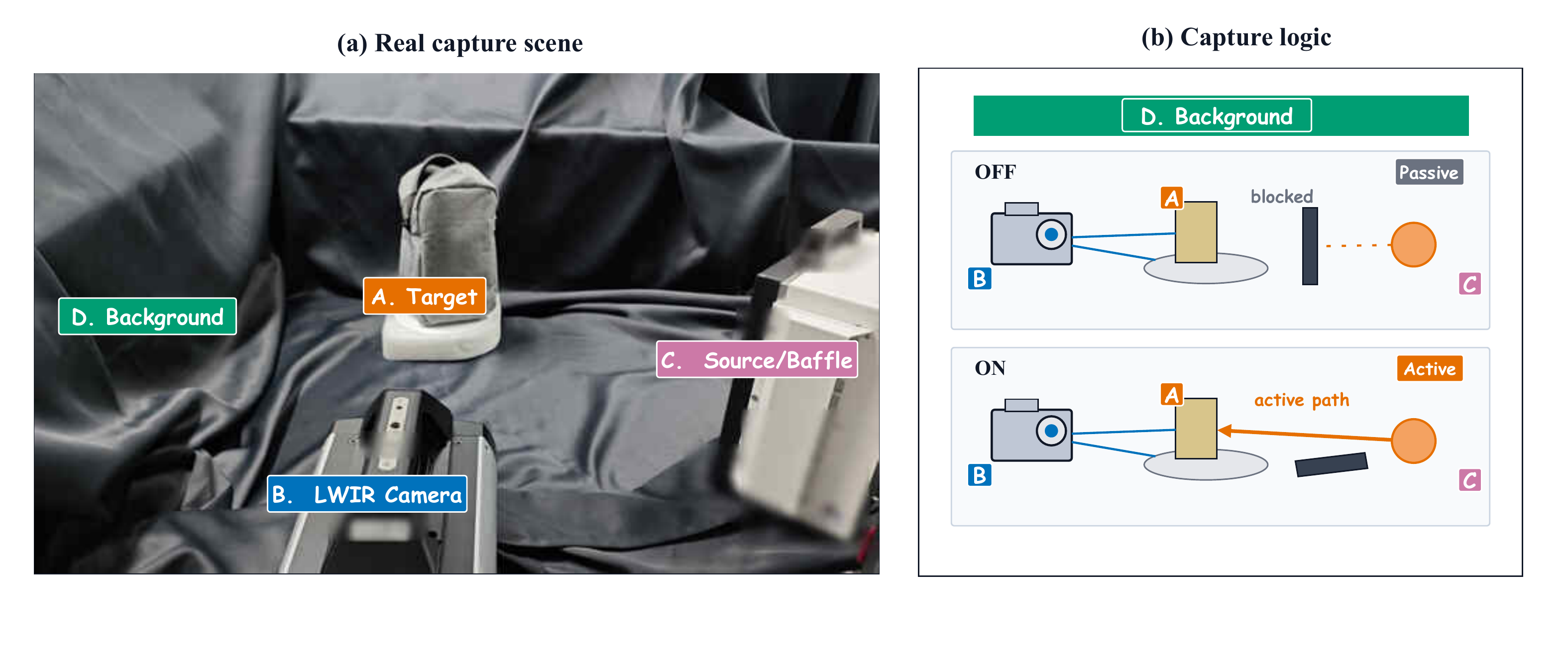}
\caption{Real capture setup and capture logic. Panel (a) labels the physical
roles used throughout the supplement: target, LWIR camera, source/baffle side,
and turntable. Panel (b) uses the same A/B/C/D colors to show the two source
states under fixed geometry: OFF closes the baffle and produces passive
candidates; ON opens the baffle and produces active thermal-radiation
enhancement. Identifying equipment tags are not legible in the submitted figure.}
\label{fig:real-setup-photo}
\end{figure*}

\paragraph{Experimental configuration.}
We simulate a closed, spatially uniform indoor scene with a spectral renderer.
The camera faces the target at a distance of \(3.6\,\mathrm{m}\). During
source-on renders, an ideal blackbody source is placed on the camera side,
\(4.5\,\mathrm{m}\) from the target, at a 71.58 degree off-axis angle relative to the
camera viewing direction. The environment is assigned
\(T=283\,\mathrm{K}\) and \(e_\nu=0.9\); target surfaces use
\(T_\alpha=303\,\mathrm{K}\) and angle-independent
\(e_{\alpha\nu}=0.9\). The active source uses
\(T=323\,\mathrm{K}\) and \(e_\nu=1.0\). We neglect atmospheric absorption
and self-emission, an approximation
specific to this short-range indoor setting.

\paragraph{Rendering procedure.}
For a visible surface element \(\alpha\), we use the thermal rendering equation
to model spectral radiance~\cite{bao2023heat,dai2026hair}:
\begin{equation}
\begin{aligned}
S_{\alpha\nu}(\tilde{\mathbf z})
&=e_{\alpha\nu}B_\nu(T_\alpha) \\
&\quad+\int_{\mathcal{S}} r_{\alpha\nu}
 (\tilde{\mathbf z},\hat{\boldsymbol\rho}_{\alpha\beta})\,
 S_{\beta\nu}(\hat{\boldsymbol\rho}_{\alpha\beta})\,
 \overline V_{\alpha\beta}\,\mathrm{d}A_\beta \\
&=e_{\alpha\nu}B_\nu(T_\alpha)
 +(1-e_{\alpha\nu})X_{\alpha\nu},
\end{aligned}
\label{eq:hre}
\end{equation}
where \(B_\nu(T)\) is Planck spectral radiance,
\begin{equation}
B_\nu(T)=\frac{2h\nu^3}{c^2}
\left[\exp\!\left(\frac{h\nu}{k_{\mathrm B}T}\right)-1\right]^{-1}.
\label{eq:planck}
\end{equation}
Here, \(\tilde{\mathbf z}\) is the direction from
\(\alpha\) to the camera, and the integral aggregates radiance from every
emitting surface element \(\beta\) in the scene. Its first term is the direct
thermal emission of \(\alpha\); its second term is the incident scene radiance
reflected by \(\alpha\). The latter is represented by \(X_{\alpha\nu}\), while
\(r_{\alpha\nu}\) is the reflectance distribution function. The corresponding
normal-dependent differential view factor is
\begin{equation}
\overline V_{\alpha\beta}=
\frac{[-\hat{\boldsymbol\rho}_{\alpha\beta}^{\top}\hat{\mathbf A}_\alpha]_+
[\hat{\boldsymbol\rho}_{\alpha\beta}^{\top}\hat{\mathbf A}_\beta]_+}
{\pi\rho_{\alpha\beta}^{2}},
\label{eq:view-factor}
\end{equation}
where \(\hat{\mathbf A}_\alpha\) and \(\hat{\mathbf A}_\beta\) are outward
surface normals, \(\hat{\boldsymbol\rho}_{\alpha\beta}\) is the unit direction
from \(\alpha\) to \(\beta\), and \(\rho_{\alpha\beta}\) is their distance.
The nonnegative clipping operator suppresses back-facing pairs; ray casting
rejects paths occluded by intervening geometry.

We estimate the integral in Eq.~\eqref{eq:hre} by normal-guided Monte Carlo
path tracing. Let \(V_{\alpha\beta}\) denote the Monte Carlo transport weight
obtained by integrating \(\overline V_{\alpha\beta}\) over the finite surface
element \(\beta\). Because the scene is closed, the initialization contains no
sky-radiance term and is determined solely by direct emission:
\begin{equation}
X^{(0)}_{\alpha\nu}=
\sum_{\beta\ne\alpha}V_{\alpha\beta}e_{\beta\nu}B_\nu(T_\beta),
\label{eq:hre-initialization}
\end{equation}
and then propagate reflected radiance through the same transport estimator:
\begin{equation}
\begin{aligned}
X^{(n+1)}_{\alpha\nu}
&=X^{(0)}_{\alpha\nu}
+\sum_{\beta\ne\alpha}V_{\alpha\beta}
\bigl(1-e_{\beta\nu}\bigr)X^{(n)}_{\beta\nu}, \\
&\hspace{7.5em} n=0,\ldots,N-1.
\end{aligned}
\label{eq:hre-iteration}
\end{equation}
Thus, each update accumulates one additional reflected bounce after the
direct-emission initialization. We use \(N=4\), which is sufficient for
numerical stabilization in our scenes, following the multi-bounce Monte Carlo
treatment.

We simulate the \(8\)--\(12\,\mu\mathrm{m}\) band and map the spectral
radiance to a single camera channel through
\begin{equation}
S_\alpha=\frac{\int_\nu R_\nu S_{\alpha\nu}\,\mathrm{d}\nu}
{\int_\nu R_\nu\,\mathrm{d}\nu},
\label{eq:camera-integration}
\end{equation}
where \(R_\nu\) is the response corresponding to a Gaussian spectral response
centered at \(10\,\mu\mathrm{m}\), with \(4\,\mu\mathrm{m}\) full width at
half maximum and support limited to \(8\)--\(12\,\mu\mathrm{m}\).

\subsection{Real LWIR Benchmark}

The real benchmark contains six independently captured sequences:
\texttt{bag}, \texttt{doll\_01}, \texttt{doll\_02}, \texttt{doll\_03},
\texttt{doll\_04}, and \texttt{doll\_05}. Each final sequence contains 180
chronologically renumbered frames. Nineteen active anchors are placed at indices
\(1,11,\ldots,171,180\). The first 17 intervals contribute nine interior
targets each and the final \((171,180)\) interval contributes eight, giving 161
targets per sequence.

Before acquisition, the FLIR X8581 thermal camera was radiometrically
calibrated against a blackbody reference. It operates over the
\(7.5\)--\(12.5~\mu\mathrm{m}\) LWIR band with a 17 mm LWIR lens matched to
that band. The camera, source/baffle assembly, and turntable base remain fixed
during each capture, while the target is carried by the turntable. In the active
state, the blackbody source is stabilized at \(130^\circ\mathrm{C}\); both the
target and indoor environment are maintained at
\(20^\circ\mathrm{C}\). A mechanical baffle switches between source-off and
source-on states without moving the optical setup.

The turntable completes one revolution in 2 minutes, corresponding to
\(3^\circ\)/s and a full 360-degree view sweep. After thermal stabilization
and non-uniformity correction, each sequence is recorded continuously.

\FloatBarrier

\section{Evaluation Protocols}
\label{sec:metrics}

We use complementary evaluation protocols according to the availability of
dense target-time texture. The synthetic benchmark provides paired
renders and therefore supports full-reference evaluation.
The real benchmark provides no dense target texture; its evaluation is
accordingly no-reference and characterizes properties of the reconstructed
images.

\subsection{Supervised Evaluation on the Synthetic Benchmark}

For a predicted texture \(\hat I\) and its target \(I\), both clamped to
\([0,1]\), we report pixel fidelity, structural similarity, interpolation
error, and boundary preservation. With \(N=HW\) pixels,
\begin{equation}
  \begin{aligned}
    &\mathrm{MSE}=\frac{1}{N}\sum_p(\hat I(p)-I(p))^2,\\
    &\mathrm{PSNR}=-10\log_{10}(\mathrm{MSE}).
  \end{aligned}
  \label{eq:psnr}
\end{equation}
PSNR measures radiometric agreement, while
SSIM~\cite{wang2004image} measures local luminance, contrast, and structural
consistency using an \(11\times11\) Gaussian window with \(\sigma=1.5\),
\(C_1=0.01^2\), and \(C_2=0.03^2\). We additionally report 8-bits interpolation
error and its normalized form:
\begin{equation}
  \begin{aligned}
    &\mathrm{IE}=\frac{1}{N}\sum_p
    \left|\mathrm{round}(255\hat I(p))-\mathrm{round}(255I(p))\right|,\\
    &\mathrm{NIE}=\frac{\mathrm{IE}}{255}.
  \end{aligned}
  \label{eq:ie-nie}
\end{equation}
Boundary preservation is measured by Edge-F1@2px. We extract Canny edge maps
from rounded 8-bit images using thresholds 100 and 200, aperture size 3, and
the \(L_2\)-gradient option. With predicted and target edge maps \(\hat E\)
and \(E\), and square radius-2 dilation \(D_2(\cdot)\),
\begin{equation}
  \begin{aligned}
  &P_e=\frac{|\hat E\cap D_2(E)|}{|\hat E|},\qquad
  R_e=\frac{|E\cap D_2(\hat E)|}{|E|},\\
  &\mathrm{Edge\mbox{-}F1@2px}=\frac{2P_eR_e}{P_e+R_e}.
  \end{aligned}
  \label{eq:edge-f1}
\end{equation}
If both edge maps are empty, Edge-F1@2px is set to 1.0. Higher PSNR, SSIM,
and Edge-F1@2px and lower IE and NIE indicate better reconstruction.

\subsection{No-Reference Evaluation on the Real Benchmark}

For each prediction, let \(P\in\{0,\ldots,255\}^{H\times W}\) denote its
8-bit grayscale representation and let \(Y=P/255\) be the corresponding
unit-range image. We compute no-reference measures of information content,
local variation, cross-anchor consistency, and perceptual quality. Entropy is
computed from the histogram \(p_b\) of \(P\):
\begin{equation}
  \mathrm{En}=-\sum_{b=0}^{255}p_b\log_2p_b.
  \label{eq:entropy}
\end{equation}
With forward differences \(\Delta_xY_{i,j}=Y_{i,j+1}-Y_{i,j}\) and
\(\Delta_yY_{i,j}=Y_{i+1,j}-Y_{i,j}\), average gradient and standard
deviation are reported in 8-bit intensity units:
\begin{align}
  \mathrm{AG} &=
  \frac{255}{(H-1)(W-1)}\sum_{i=1}^{H-1}\sum_{j=1}^{W-1}
  \sqrt{\frac{\Delta_xY_{i,j}^2+\Delta_yY_{i,j}^2}{2}},\\
  \mathrm{SD} &=
  255\sqrt{\frac{1}{N}\sum_p(Y(p)-\bar Y)^2},
  \qquad
  \bar Y=\frac{1}{N}\sum_pY(p).
  \label{eq:ag-sd}
\end{align}
Let \(A\) and \(B\) be the left and right active anchors after common
grayscale normalization. The sum of correlations of differences
(SCD)~\cite{aslantas2015scd} is
\begin{equation}
  \mathrm{SCD}=\rho(P-A,B)+\rho(P-B,A),
  \label{eq:scd}
\end{equation}
where \(\rho(\cdot,\cdot)\) is Pearson correlation over pixels. Finally, the
perceptual index combines NIQE~\cite{mittal2013making,li2026unim} and
NRQM~\cite{ma2017learning} in the standard PIRM form~\cite{blau2018pirm}:
\begin{equation}
  \mathrm{PI}=\frac{1}{2}\left(\mathrm{NIQE}+10-\mathrm{NRQM}\right).
  \label{eq:pi}
\end{equation}
Higher En, AG, SD, and SCD indicate greater information content, local
variation, or cross-anchor consistency; lower NIQE and PI indicate better
perceptual quality. 

\section{Training and Inference Details}
\subsection{End-to-End Inference}
\label{sec:inference}

\subsubsection{Mixed Active--Passive Observations}

Consider a sequence of \(N\) thermal frames. Illumination is enabled only at
sparse active keyframes \(\mathcal K\subset\{1,\ldots,N\}\), yielding the
observations \(\mathcal S^{\mathrm{on}}=\{S_k^{\mathrm{on}}\}_{k\in\mathcal K}\).
All remaining frames are acquired passively and form
\(\mathcal S^{\mathrm{off}}=\{S_i^{\mathrm{off}}\}_{i\in\overline{\mathcal K}}\),
where \(\overline{\mathcal K}=\{1,\ldots,N\}\setminus\mathcal K\). The
source-off state at an active keyframe is therefore unobserved. From these mixed
observations, T$^2$exture recovers a dense texture
sequence \(\hat{\mathcal X}=\{\hat X_t\}_{t=1}^{N}\). It uses pretrained
AMT-L for Stage~1, denoted by \(f_\theta\), and the adapted AMT-L model for
Stage~2, denoted by \(g_\phi\). Model-scale variants are specified in
Section~\ref{sec:training-protocol}.

\subsubsection{Stage~1: Source-off Passive State Estimation}

For each interior active keyframe \(k\), Stage~1 estimates the missing
source-off state from its adjacent passive observations:
\begin{equation}
  \widehat S_k^{\mathrm{off}}=
  f_\theta\!\left(S_{k-1}^{\mathrm{off}},S_{k+1}^{\mathrm{off}},
  t=\tfrac{1}{2}\right).
  \label{eq:stage1-source-off}
\end{equation}
The active observation and its estimate form a texture anchor,
\begin{equation}
  X_k=\left[S_k^{\mathrm{on}}-\widehat S_k^{\mathrm{off}}\right]_+,
  \label{eq:supp-texture-anchor}
\end{equation}
where \([u]_+=\max(u,0)\). Stage~1 thereby completes the passive sequence:
\begin{equation}
  \widetilde S_i^{\mathrm{off}}=
  \begin{cases}
    S_i^{\mathrm{off}}, & i\in\overline{\mathcal K},\\
    \widehat S_i^{\mathrm{off}}, & i\in\mathcal K.
  \end{cases}
  \label{eq:completed-passive-sequence}
\end{equation}

\subsubsection{Stage~2: Structure- and Semantic-Guided Texture Propagation}

Let \(X_0\) and \(X_1\) be the texture anchors bracketing a target time
\(t\in(0,1)\). We use the centered passive context from the completed passive sequence
with radius \(r\):
\begin{equation}
  \begin{aligned}
  \mathcal C_t={}&\Big(
  (\widetilde S_{\tau_i^-}^{\mathrm{off}})_{i=r}^{1},
  \widetilde S_t^{\mathrm{off}},
  (\widetilde S_{\tau_i^+}^{\mathrm{off}})_{i=1}^{r}\Big).
  \end{aligned}
  \label{eq:supp-passive-context}
\end{equation}
Here, \(t\) is supplied directly as the normalized location between its two
active anchors. Stage~2 predicts the target texture as
\begin{equation}
  \widehat X_t=g_\phi(X_0,X_1,t,\mathcal C_t).
  \label{eq:stage2-inference}
\end{equation}
Applying this operation to every valid target time and retaining the Stage~1
anchors at active keyframes yields \(\hat{\mathcal X}\).

\begin{algorithm}[t]
\caption{Texture Inference from Mixed Observations}
\label{alg:end-to-end-inference}
\begin{algorithmic}[1]
\REQUIRE \(\mathcal S^{\mathrm{on}}\), \(\mathcal S^{\mathrm{off}}\), \(\mathcal K\),
\(f_\theta\), \(g_\phi\)
\ENSURE Dense texture sequence \(\hat{\mathcal X}=\{\widehat X_t\}_{t=1}^{N}\)
\STATE \textbf{Stage 1: Source-off Passive State Estimation}
\FOR{each interior \(k\in\mathcal K\)}
  \STATE \(\widehat S_k^{\mathrm{off}}\leftarrow
  f_\theta(S_{k-1}^{\mathrm{off}},S_{k+1}^{\mathrm{off}},\tfrac{1}{2})\)
  \STATE \(X_k\leftarrow[S_k^{\mathrm{on}}-\widehat S_k^{\mathrm{off}}]_+\),
  \quad \(\hat X_k\leftarrow X_k\)
\ENDFOR
\STATE \(\widetilde S_i^{\mathrm{off}}\leftarrow
\begin{cases}S_i^{\mathrm{off}},&i\in\overline{\mathcal K},\\
\widehat S_i^{\mathrm{off}},&i\in\mathcal K\end{cases}\)
\STATE \textbf{Stage 2: Structure- and Semantic-Guided Texture Propagation}
\FOR{each target time \(t\) between adjacent anchors \(X_0,X_1\)}
  \STATE \(\mathcal C_t\leftarrow
  \big((\widetilde S_{\tau_i^-}^{\mathrm{off}})_{i=r}^{1},
  \widetilde S_t^{\mathrm{off}},
  (\widetilde S_{\tau_i^+}^{\mathrm{off}})_{i=1}^{r}\big)\)
  \STATE \(\hat X_t\leftarrow g_\phi(X_0,X_1,t,\mathcal C_t)\)
\ENDFOR
\STATE \(\hat{\mathcal X}\leftarrow
\{\hat X_k=X_k\}_{k\in\mathcal K}\cup\{\hat X_t\}_{t\in\overline{\mathcal K}}\)
\STATE \RETURN \(\hat{\mathcal X}\)
\end{algorithmic}
\end{algorithm}

\subsection{Training Details}
\label{sec:training-protocol}

All models are trained exclusively on the synthetic training split. Validation
and checkpoint selection use the synthetic validation split, and the selected
model is evaluated on the synthetic test split. Table~\ref{tab:training-protocol}
summarizes the default training configuration; ablations vary only the setting
under study.

\begin{table}[t]
\centering
\small
\setlength{\tabcolsep}{4pt}
\begin{tabular}{>{\raggedright\arraybackslash}p{0.34\linewidth}>{\raggedright\arraybackslash}p{0.58\linewidth}}
\toprule
Setting & Default value \\
\midrule
Backbone & Pretrained AMT-S/L/G. \\
Active stride & 10 frames. \\
Passive context & Five frames (\(r=2\)) in Eq.~\eqref{eq:supp-passive-context}. \\
Pseudo-flow targets & Target-to-left/right-anchor flow fields. \\
Training crop & \(384\times384\) pixels. \\
Augmentation & None. \\
Batch size & Training: 4; validation/testing: 1. \\
Schedule & 10,000 adaptation iterations, then 5,000 fine-tuning iterations. \\
Learning rates & Adaptation: \(2{\times}10^{-4}\); Fine-tuning:\(5{\times}10^{-5}\). \\
Optimizer & AdamW, \((\beta_1,\beta_2)=(0.9,0.99)\), weight decay \(10^{-5}\). \\
Layer-wise LR decay & 0.8 during fine-tuning. \\
Checkpoint selection & Every 500 iterations; highest PSNR. \\
Random seed & 2026. \\
\bottomrule
\end{tabular}
\caption{Default training configuration.}
\label{tab:training-protocol}
\end{table}

\subsubsection{Training Loss}

Both Stage~2 adaptation and fine-tuning optimize the same composite objective,
which combines pixel-wise reconstruction, local structural consistency, and
pseudo-flow regularization:
\begin{equation}
  \mathcal{L} =
  1.0\mathcal{L}_{\mathrm{char}}
  +0.1\mathcal{L}_{\mathrm{css}}
  +0.001\mathcal{L}_{\mathrm{flow}}.
  \label{eq:loss-total}
\end{equation}
The Charbonnier loss term~\cite{charbonnier1994deterministic} is
\begin{equation}
  \mathcal{L}_{\mathrm{char}}
  =
  \frac{1}{HW}
  \sum_{p\in\Omega}
  \sqrt{(\xhat_t(p)-\texture_t(p))^2+10^{-6}} .
  \label{eq:charbonnier}
\end{equation}

For the bidirectional census loss~\cite{meister2018unflow}, define a
\(7\times7\) offset set
\(\mathcal{R}\), local difference \(\Delta_r I(p)=I(p+r)-I(p)\), and
normalized difference
\begin{equation}
  \phi_r(I,p)=
  \frac{\Delta_r I(p)}{\sqrt{0.81+\Delta_r I(p)^2}} .
\end{equation}
The structural loss is averaged over all output pixels:
\begin{equation}
  \begin{aligned}
    &\mathcal{L}_{\mathrm{css}} =
    \frac{1}{HW|\mathcal{R}|}
    \sum_{p\in\Omega}\sum_{r\in\mathcal{R}}
    \psi(d_{r,p}), \\
    &\psi(z)=\frac{z^2}{0.1+z^2},\quad
    d_{r,p}=
    \phi_r(\xhat_t,p)-\phi_r(\texture_t,p).
  \end{aligned}
  \label{eq:css}
\end{equation}

Following AMT~\cite{li2023amt}, the pseudo-flow term regularizes the
coarse-scale target-to-anchor motion. Let \(F^d\) be the stored pseudo-flow for direction
\(d\in\{0,1\}\). At the finest level, AMT predicts \(M\) flow candidates
\(\{\hat F_0^{d,m}\}_{m=1}^M\); their maximum endpoint error defines a confidence weight:
\begin{equation}
  \begin{aligned}
    &\omega^d(p)=\exp[-0.3e^d(p)], \\
    &e^d(p)=\max_m\|\hat F^{d,m}_0(p)-F^d(p)\|_2 .
  \end{aligned}
\end{equation}
With
\(\epsilon(w)=10^{-(10w-1)/3}\) and
\(\ell_{\mathrm{ada}}(z;w)=(z^2+\epsilon(w)^2)^{w/2}\), let
\(\hat F_\ell^d\) denote the single flow predicted at each remaining pyramid
level \(\ell\geq1\), and define
\begin{equation}
  \delta_{\ell}^{d,c}(p)=
  [\mathcal{U}_{2^\ell}(\hat F_\ell^d)]_c(p)-F^d_c(p).
  \label{eq:flow-discrepancy}
\end{equation}
The flow loss supervises these coarse levels:
\begin{equation}
  \mathcal{L}_{\mathrm{flow}}=
  \sum_{d\in\{0,1\}}\sum_{\ell=1}^{L-1}
  \operatorname*{mean}_{p,c}
  \ell_{\mathrm{ada}}\!\left(\delta_{\ell}^{d,c}(p);\omega^d(p)\right).
  \label{eq:flow-loss}
\end{equation}
where \(\mathcal{U}_{2^\ell}\) is bilinear resizing with the
flow-vector scaling.

\subsubsection{Model Adaptation Details}

The AMT backbone~\cite{li2023amt} is designed for RGB frame interpolation,
whereas our texture signal is a single-channel thermal residual. An input
adapter maps the texture anchors to the backbone input channels and is
initialized to preserve channel replication. The RGB-style output is averaged
back to one texture channel.

Passive contexts are stacked along the channel dimension and encoded by a
three-level passive branch. At each level, a zero-initialized \(1\times1\)
convolution injects a residual feature into the corresponding AMT decoder
feature. This initialization preserves the pretrained prediction at the start
of adaptation, allowing the passive branch to learn residual corrections.

Training proceeds in two phases. During adaptation, the pretrained AMT backbone
is frozen and only the texture adapter, time embedding, passive encoder, and
zero-initialized residual modules are optimized. During fine-tuning, these
modules remain trainable and the late AMT refinement modules are unfrozen with
layer-wise learning-rate decay. This procedure is shared by AMT-S, AMT-L, and
AMT-G.

\FloatBarrier
\bibliography{references}

\begin{thebibliography}{52}
\providecommand{\natexlab}[1]{#1}

\bibitem[{Arbel{\'a}ez et~al.(2011)Arbel{\'a}ez, Maire, Fowlkes, and Malik}]{arbelaez2011contour}
Arbel{\'a}ez, P.; Maire, M.; Fowlkes, C.; and Malik, J. 2011.
\newblock Contour Detection and Hierarchical Image Segmentation.
\newblock \emph{IEEE Transactions on Pattern Analysis and Machine Intelligence}, 33(5): 898--916.

\bibitem[{Aslantas and Bendes(2015)}]{aslantas2015scd}
Aslantas, V.; and Bendes, E. 2015.
\newblock A New Image Quality Metric for Image Fusion: The Sum of the Correlations of Differences.
\newblock \emph{AEU -- International Journal of Electronics and Communications}, 69(12): 1890--1896.

\bibitem[{Aveni et~al.(2024)Aveni, Laiolo, Campus, Massimetti, and Coppola}]{aveni2024tirvolch}
Aveni, S.; Laiolo, M.; Campus, A.; Massimetti, F.; and Coppola, D. 2024.
\newblock {TIRVolcH}: Thermal Infrared Recognition of Volcanic Hotspots: A Single-Band {TIR}-Based Algorithm to Detect Low-to-High Thermal Anomalies in Volcanic Regions.
\newblock \emph{Remote Sensing of Environment}, 315: 114388.

\bibitem[{Bao et~al.(2024)Bao, Jape, Schramka, Wang, McGraw, and Jacob}]{bao2024blurry}
Bao, F.; Jape, S.; Schramka, A.; Wang, J.; McGraw, T.~E.; and Jacob, Z. 2024.
\newblock Why thermal images are blurry.
\newblock \emph{Optics Express}, 32(3): 3852--3865.

\bibitem[{Bao et~al.(2023)Bao, Wang, Sureshbabu, Sreekumar, Yang, Aggarwal, Boddeti, and Jacob}]{bao2023heat}
Bao, F.; Wang, X.; Sureshbabu, S.~H.; Sreekumar, G.; Yang, L.; Aggarwal, V.; Boddeti, V.~N.; and Jacob, Z. 2023.
\newblock Heat-assisted detection and ranging.
\newblock \emph{Nature}, 619(7971): 743--748.

\bibitem[{Blau et~al.(2018)Blau, Mechrez, Timofte, Michaeli, and Zelnik-Manor}]{blau2018pirm}
Blau, Y.; Mechrez, R.; Timofte, R.; Michaeli, T.; and Zelnik-Manor, L. 2018.
\newblock The 2018 PIRM Challenge on Perceptual Image Super-Resolution.
\newblock In \emph{Proceedings of the European Conference on Computer Vision Workshops}.

\bibitem[{Canny(1986)}]{canny1986computational}
Canny, J. 1986.
\newblock A Computational Approach to Edge Detection.
\newblock \emph{IEEE Transactions on Pattern Analysis and Machine Intelligence}, PAMI-8(6): 679--698.

\bibitem[{Charbonnier et~al.(1994)Charbonnier, Blanc-F{\'e}raud, Aubert, and Barlaud}]{charbonnier1994deterministic}
Charbonnier, P.; Blanc-F{\'e}raud, L.; Aubert, G.; and Barlaud, M. 1994.
\newblock Two Deterministic Half-Quadratic Regularization Algorithms for Computed Imaging.
\newblock In \emph{Proceedings of the IEEE International Conference on Image Processing}, volume~2, 168--172.

\bibitem[{Dai et~al.(2026{\natexlab{a}})Dai, Lin, Song, Chen, Chen, Yuan, and Bao}]{dai2026hair}
Dai, C.; Lin, J.; Song, B.; Chen, Y.; Chen, J.; Yuan, X.; and Bao, F. 2026{\natexlab{a}}.
\newblock {HADAR}-Based Thermal Infrared Hyperspectral Image Restoration.
\newblock arXiv:2605.13664.

\bibitem[{Dai et~al.(2026{\natexlab{b}})Dai, Lin, Xu, Song, Xie, and Bao}]{dai2026tex1500}
Dai, C.; Lin, J.; Xu, H.; Song, B.; Xie, Z.; and Bao, F. 2026{\natexlab{b}}.
\newblock {TeX}-1500: A Paired Real-World {LWIR} Hyperspectral Dataset and Benchmark for Temperature-Emissivity-Texture Decomposition.
\newblock arXiv:2606.03806.

\bibitem[{Dorken~Gallastegi et~al.(2025)Dorken~Gallastegi, Rueda-Chac{\'o}n, Stevens, and Goyal}]{gallastegi2025absorption}
Dorken~Gallastegi, U.; Rueda-Chac{\'o}n, H.; Stevens, M.~J.; and Goyal, V.~K. 2025.
\newblock Absorption-Based, Passive Range Imaging from Hyperspectral Thermal Measurements.
\newblock \emph{IEEE Transactions on Pattern Analysis and Machine Intelligence}, 47(5): 4044--4060.

\bibitem[{Dorken~Gallastegi et~al.(2026)Dorken~Gallastegi, Shangguan, Choudhary, Agarwal, Rueda-Chac{\'o}n, Stevens, and Goyal}]{gallastegi2026ozone}
Dorken~Gallastegi, U.; Shangguan, W.; Choudhary, V.; Agarwal, A.; Rueda-Chac{\'o}n, H.; Stevens, M.~J.; and Goyal, V.~K. 2026.
\newblock Ozone Cues Mitigate Reflected Downwelling Radiance in LWIR Absorption-Based Ranging.
\newblock \emph{IEEE Transactions on Computational Imaging}, 12: 587--600.

\bibitem[{Erdozain et~al.(2020)Erdozain, Ichimaru, Maeda, Kawasaki, Raskar, and Kadambi}]{erdozain2020structured}
Erdozain, J.; Ichimaru, K.; Maeda, T.; Kawasaki, H.; Raskar, R.; and Kadambi, A. 2020.
\newblock {3D} Imaging for Thermal Cameras Using Structured Light.
\newblock In \emph{2020 IEEE International Conference on Image Processing}, 2795--2799.

\bibitem[{Guo, Li, and Loy(2024)}]{guo2024gimmvfi}
Guo, Z.; Li, W.; and Loy, C.~C. 2024.
\newblock Generalizable Implicit Motion Modeling for Video Frame Interpolation.
\newblock In \emph{Advances in Neural Information Processing Systems}, volume~37, 63747--63770.

\bibitem[{Hai et~al.(2025)Hai, Wang, Su, Jiang, and Hu}]{hai2025hfd}
Hai, Y.; Wang, G.; Su, T.; Jiang, W.; and Hu, Y. 2025.
\newblock Hierarchical Flow Diffusion for Efficient Frame Interpolation.
\newblock In \emph{Proceedings of the IEEE/CVF Conference on Computer Vision and Pattern Recognition}, 22943--22952.

\bibitem[{Han et~al.(2025)Han, Zheng, Ling, and Jia}]{han2025phasor}
Han, D.; Zheng, C.; Ling, Z.; and Jia, S. 2025.
\newblock Hyperspectral Phasor Thermography.
\newblock \emph{Cell Reports Physical Science}, 6(3): 102501.

\bibitem[{Hu, Hu, and Chen(2024)}]{hu2024tesr}
Hu, L.; Hu, L.; and Chen, M. 2024.
\newblock Edge-Enhanced Infrared Image Super-Resolution Reconstruction Model Under Transformer.
\newblock \emph{Scientific Reports}, 14: 15585.

\bibitem[{{International Commission on Non-Ionizing Radiation Protection}(2013)}]{icnirp2013incoherent}
{International Commission on Non-Ionizing Radiation Protection}. 2013.
\newblock {ICNIRP} Guidelines on Limits of Exposure to Incoherent Visible and Infrared Radiation.
\newblock \emph{Health Physics}, 105(1): 74--96.

\bibitem[{Jiang et~al.(2021)Jiang, Campbell, Lu, Li, and Hartley}]{jiang2021gma}
Jiang, S.; Campbell, D.; Lu, Y.; Li, H.; and Hartley, R. 2021.
\newblock Learning To Estimate Hidden Motions With Global Motion Aggregation.
\newblock In \emph{Proceedings of the IEEE/CVF International Conference on Computer Vision}, 9772--9781.

\bibitem[{Kong et~al.(2022)Kong, Jiang, Luo, Chu, Huang, Tai, Wang, and Yang}]{kong2022ifrnet}
Kong, L.; Jiang, B.; Luo, D.; Chu, W.; Huang, X.; Tai, Y.; Wang, C.; and Yang, J. 2022.
\newblock {IFRNet}: Intermediate Feature Refine Network for Efficient Frame Interpolation.
\newblock In \emph{Proceedings of the IEEE/CVF Conference on Computer Vision and Pattern Recognition}, 1969--1978.

\bibitem[{Landmann et~al.(2021)Landmann, Speck, Dietrich, Heist, K{\"u}hmstedt, T{\"u}nnermann, and Notni}]{landmann2021thermal}
Landmann, M.; Speck, H.; Dietrich, P.; Heist, S.; K{\"u}hmstedt, P.; T{\"u}nnermann, A.; and Notni, G. 2021.
\newblock High-Resolution Sequential Thermal Fringe Projection Technique for Fast and Accurate {3D} Shape Measurement of Transparent Objects.
\newblock \emph{Applied Optics}, 60(8): 2362--2371.

\bibitem[{Li et~al.(2026)Li, Guo, Zhang, Zhang, Zhao, Li, Zhou, Zheng, Yan, Wu et~al.}]{li2026unim}
Li, Y.; Guo, M.; Zhang, K.; Zhang, S.; Zhao, Y.; Li, H.; Zhou, C.; Zheng, W.; Yan, Y.; Wu, S.; et~al. 2026.
\newblock UniM: A Unified Any-to-Any Interleaved Multimodal Benchmark.
\newblock \emph{arXiv preprint arXiv:2603.05075}.

\bibitem[{Li et~al.(2023)Li, Zhu, Han, Hou, Guo, and Cheng}]{li2023amt}
Li, Z.; Zhu, Z.-L.; Han, L.-H.; Hou, Q.; Guo, C.-L.; and Cheng, M.-M. 2023.
\newblock {AMT}: All-Pairs Multi-Field Transforms for Efficient Frame Interpolation.
\newblock In \emph{Proceedings of the IEEE/CVF Conference on Computer Vision and Pattern Recognition}, 9801--9810.

\bibitem[{Liu et~al.(2019)Liu, Sui, Kuang, Liu, Gu, and Chen}]{liu2019adaptive}
Liu, C.; Sui, X.; Kuang, X.; Liu, Y.; Gu, G.; and Chen, Q. 2019.
\newblock Adaptive Contrast Enhancement for Infrared Images Based on the Neighborhood Conditional Histogram.
\newblock \emph{Remote Sensing}, 11(11): 1381.

\bibitem[{Liu et~al.(2024{\natexlab{a}})Liu, Zhang, Zhao, and Wang}]{liu2024sgmvfi}
Liu, C.; Zhang, G.; Zhao, R.; and Wang, L. 2024{\natexlab{a}}.
\newblock Sparse Global Matching for Video Frame Interpolation with Large Motion.
\newblock In \emph{Proceedings of the IEEE/CVF Conference on Computer Vision and Pattern Recognition}, 19125--19134.

\bibitem[{Liu et~al.(2024{\natexlab{b}})Liu, Zhu, Jin, and Huang}]{liu2024radicular}
Liu, H.; Zhu, Z.; Jin, X.; and Huang, P. 2024{\natexlab{b}}.
\newblock The Diagnostic Accuracy of Infrared Thermography in Lumbosacral Radicular Pain: A Prospective Study.
\newblock \emph{Journal of Orthopaedic Surgery and Research}, 19(1): 409.

\bibitem[{Liu et~al.(2026)Liu, Fan, Chen, Huang, and Zhang}]{liu2026ader}
Liu, S.; Fan, C.; Chen, Z.; Huang, X.; and Zhang, L. 2026.
\newblock Absorption-Feature-Guided Distance-Decoupled Estimation and Band Selection for {LWIR} Hyperspectral Passive Ranging.
\newblock arXiv:2606.31824.

\bibitem[{Lu, Zhang, and Yin(2025)}]{defusion2025}
Lu, Q.; Zhang, H.; and Yin, L. 2025.
\newblock Infrared and Visible Image Fusion via Dual Encoder Based on Dense Connection.
\newblock \emph{Pattern Recognition}, 163: 111476.

\bibitem[{Lyu and Chen(2025)}]{lyu2025tlbvfi}
Lyu, Z.; and Chen, C. 2025.
\newblock {TLB-VFI}: Temporal-Aware Latent Brownian Bridge Diffusion for Video Frame Interpolation.
\newblock In \emph{Proceedings of the IEEE/CVF International Conference on Computer Vision}, 16260--16269.

\bibitem[{Ma et~al.(2017)Ma, Yang, Yang, and Yang}]{ma2017learning}
Ma, C.; Yang, C.-Y.; Yang, X.; and Yang, M.-H. 2017.
\newblock Learning a No-Reference Quality Metric for Single-Image Super-Resolution.
\newblock \emph{Computer Vision and Image Understanding}, 158: 1--16.

\bibitem[{Meister, Hur, and Roth(2018)}]{meister2018unflow}
Meister, S.; Hur, J.; and Roth, S. 2018.
\newblock {UnFlow}: Unsupervised Learning of Optical Flow With a Bidirectional Census Loss.
\newblock In \emph{Proceedings of the AAAI Conference on Artificial Intelligence}, volume~32, 7251--7259.

\bibitem[{Mittal, Soundararajan, and Bovik(2013)}]{mittal2013making}
Mittal, A.; Soundararajan, R.; and Bovik, A.~C. 2013.
\newblock Making a Completely Blind Image Quality Analyzer.
\newblock \emph{IEEE Signal Processing Letters}, 20(3): 209--212.

\bibitem[{Mou et~al.(2024)Mou, Wang, Xie, Wu, Zhang, Qi, and Shan}]{mou2024t2iadapter}
Mou, C.; Wang, X.; Xie, L.; Wu, Y.; Zhang, J.; Qi, Z.; and Shan, Y. 2024.
\newblock {T2I}-Adapter: Learning Adapters to Dig Out More Controllable Ability for Text-to-Image Diffusion Models.
\newblock In \emph{Proceedings of the AAAI Conference on Artificial Intelligence}, volume~38, 4296--4304.

\bibitem[{Ng et~al.(2024)Ng, Dhruval, Shalabi, Jape, Wang, and Jacob}]{ng2024thermalvoyager}
Ng, A.; Dhruval, P.; Shalabi, J.; Jape, S.; Wang, X.; and Jacob, Z. 2024.
\newblock Thermal Voyager: A Comparative Study of {RGB} and Thermal Cameras for Night-Time Autonomous Navigation.
\newblock In \emph{Proceedings of the IEEE International Conference on Robotics and Automation}, 14116--14122.

\bibitem[{Niklaus, Mai, and Liu(2017)}]{niklaus2017sepconv}
Niklaus, S.; Mai, L.; and Liu, F. 2017.
\newblock Video Frame Interpolation via Adaptive Separable Convolution.
\newblock In \emph{Proceedings of the IEEE International Conference on Computer Vision}, 261--270.

\bibitem[{Peng et~al.(2026)Peng, Li, Huang, Zheng, Wang, Chen, Dai, Li, Zou, and Xiong}]{peng2026ldfvfi}
Peng, X.; Li, H.; Huang, Y.; Zheng, Z.; Wang, Y.; Chen, X.; Dai, W.; Li, C.; Zou, J.; and Xiong, H. 2026.
\newblock Towards Holistic Modeling for Video Frame Interpolation with Auto-Regressive Diffusion Transformers.
\newblock In \emph{Proceedings of the IEEE/CVF Conference on Computer Vision and Pattern Recognition}, 11448--11458.

\bibitem[{Seo, Oh, and Kim(2025)}]{seo2025bimvfi}
Seo, W.; Oh, J.; and Kim, M. 2025.
\newblock {BiM-VFI}: Bidirectional Motion Field-Guided Frame Interpolation for Video with Non-Uniform Motions.
\newblock In \emph{Proceedings of the IEEE/CVF Conference on Computer Vision and Pattern Recognition}, 7244--7253.

\bibitem[{Sheinin, Sankaranarayanan, and Narasimhan(2024)}]{sheinin2024thermal}
Sheinin, M.; Sankaranarayanan, A.~C.; and Narasimhan, S.~G. 2024.
\newblock Projecting Trackable Thermal Patterns for Dynamic Computer Vision.
\newblock In \emph{Proceedings of the IEEE/CVF Conference on Computer Vision and Pattern Recognition}, 25223--25232.

\bibitem[{Speck et~al.(2026)Speck, Landmann, Ramm, Heist, K{\"u}hmstedt, and Notni}]{speck2026thermal}
Speck, H.; Landmann, M.; Ramm, R.; Heist, S.; K{\"u}hmstedt, P.; and Notni, G. 2026.
\newblock Analysis of the Measurement Accuracy of a Thermal {3D} Sensor for Transparent Objects.
\newblock \emph{Measurement}, 258: 119068.

\bibitem[{Tancik et~al.(2020)Tancik, Srinivasan, Mildenhall, Fridovich-Keil, Raghavan, Singhal, Ramamoorthi, Barron, and Ng}]{tancik2020fourier}
Tancik, M.; Srinivasan, P.~P.; Mildenhall, B.; Fridovich-Keil, S.; Raghavan, N.; Singhal, U.; Ramamoorthi, R.; Barron, J.~T.; and Ng, R. 2020.
\newblock Fourier Features Let Networks Learn High Frequency Functions in Low Dimensional Domains.
\newblock In \emph{Advances in Neural Information Processing Systems}, volume~33.

\bibitem[{Tang, Li, and Ma(2026)}]{maskdifuser2026}
Tang, L.; Li, C.; and Ma, J. 2026.
\newblock Mask-DiFuser: A Masked Diffusion Model for Unified Unsupervised Image Fusion.
\newblock \emph{IEEE Transactions on Pattern Analysis and Machine Intelligence}, 48(1): 591--608.

\bibitem[{Teed and Deng(2020)}]{teed2020raft}
Teed, Z.; and Deng, J. 2020.
\newblock {RAFT}: Recurrent All-Pairs Field Transforms for Optical Flow.
\newblock In \emph{European Conference on Computer Vision}, 402--419.

\bibitem[{Teng et~al.(2024)Teng, Ren, Hu, and Dou}]{teng2024sdgsat1}
Teng, Y.; Ren, H.; Hu, Y.; and Dou, C. 2024.
\newblock Land Surface Temperature Retrieval from {SDGSAT-1} Thermal Infrared Spectrometer Images: Algorithm and Validation.
\newblock \emph{Remote Sensing of Environment}, 315: 114412.

\bibitem[{Wang et~al.(2024)Wang, Qu, Zhang, and Xie}]{wang2024multifocus}
Wang, J.; Qu, H.; Zhang, Z.; and Xie, M. 2024.
\newblock New Insights into Multi-Focus Image Fusion: A Fusion Method Based on Multi-Dictionary Linear Sparse Representation and Region Fusion Model.
\newblock \emph{Information Fusion}, 105: 102230.

\bibitem[{Wang et~al.(2004)Wang, Bovik, Sheikh, and Simoncelli}]{wang2004image}
Wang, Z.; Bovik, A.~C.; Sheikh, H.~R.; and Simoncelli, E.~P. 2004.
\newblock Image Quality Assessment: From Error Visibility to Structural Similarity.
\newblock \emph{IEEE Transactions on Image Processing}, 13(4): 600--612.

\bibitem[{Xu et~al.(2026)Xu, Wang, Zhao, Chen, Lin, Cao, Zhong, She, and Bao}]{xu2026tag}
Xu, H.; Wang, D.; Zhao, C.; Chen, J.; Lin, J.; Cao, L.; Zhong, Y.; She, Y.; and Bao, F. 2026.
\newblock Universal Computational Thermal Imaging Overcoming the Ghosting Effect.
\newblock arXiv:2604.01542.

\bibitem[{Zhang et~al.(2024)Zhang, Liu, Yang, Huang, and Huang}]{zhang2024trafficnight}
Zhang, G.; Liu, Y.; Yang, X.; Huang, H.; and Huang, C. 2024.
\newblock TrafficNight: An Aerial Multimodal Benchmark for Nighttime Vehicle Surveillance.
\newblock In \emph{European Conference on Computer Vision}, 36--48.

\bibitem[{Zhang et~al.(2025)Zhang, Chen, Zhao, Lu, Fu, Xu, and Wu}]{zhang2025eden}
Zhang, Z.; Chen, H.; Zhao, H.; Lu, G.; Fu, Y.; Xu, H.; and Wu, Z. 2025.
\newblock {EDEN}: Enhanced Diffusion for High-Quality Large-Motion Video Frame Interpolation.
\newblock In \emph{Proceedings of the IEEE/CVF Conference on Computer Vision and Pattern Recognition}, 2105--2115.

\bibitem[{Zhao et~al.(2023{\natexlab{a}})Zhao, Xie, Zhao, He, and Lu}]{metafusion2023}
Zhao, W.; Xie, S.; Zhao, F.; He, Y.; and Lu, H. 2023{\natexlab{a}}.
\newblock MetaFusion: Infrared and Visible Image Fusion via Meta-Feature Embedding from Object Detection.
\newblock In \emph{Proceedings of the IEEE/CVF Conference on Computer Vision and Pattern Recognition}, 13955--13965.

\bibitem[{Zhao et~al.(2023{\natexlab{b}})Zhao, Bai, Zhang, Zhang, Xu, Lin, Timofte, and Van~Gool}]{cddfuse2023}
Zhao, Z.; Bai, H.; Zhang, J.; Zhang, Y.; Xu, S.; Lin, Z.; Timofte, R.; and Van~Gool, L. 2023{\natexlab{b}}.
\newblock CDDFuse: Correlation-Driven Dual-Branch Feature Decomposition for Multi-Modality Image Fusion.
\newblock In \emph{Proceedings of the IEEE/CVF Conference on Computer Vision and Pattern Recognition}, 5906--5916.

\bibitem[{Zhao et~al.(2023{\natexlab{c}})Zhao, Bai, Zhu, Zhang, Xu, Zhang, Zhang, Meng, Timofte, and Van~Gool}]{ddfm2023}
Zhao, Z.; Bai, H.; Zhu, Y.; Zhang, J.; Xu, S.; Zhang, Y.; Zhang, K.; Meng, D.; Timofte, R.; and Van~Gool, L. 2023{\natexlab{c}}.
\newblock DDFM: Denoising Diffusion Model for Multi-Modality Image Fusion.
\newblock In \emph{Proceedings of the IEEE/CVF International Conference on Computer Vision}, 8082--8093.

\bibitem[{Zuiderveld(1994)}]{zuiderveld1994clahe}
Zuiderveld, K.~J. 1994.
\newblock Contrast Limited Adaptive Histogram Equalization.
\newblock In Heckbert, P.~S., ed., \emph{Graphics Gems IV}, 474--485. Academic Press.

\end{thebibliography}

\end{document}